# A comprehensive overview of deep learning techniques for 3D point cloud classification and semantic segmentation

Sushmita Sarker[1] · Prithul Sarker[1] · Gunner Stone[1] · Ryan Gorman[1] · Alireza Tavakkoli[1] · George Bebis[1] · Javad Sattarvand[2]

**Abstract**
Point cloud analysis has a wide range of applications in many areas such as computer vision, robotic manipulation, and autonomous driving. While deep learning has achieved remarkable success on image-based tasks, there are many unique challenges faced by deep neural networks in processing massive, unordered, irregular and noisy 3D points. To stimulate future research, this paper analyzes recent progress in deep learning methods employed for point cloud processing and presents challenges and potential directions to advance this field. It serves as a comprehensive review on two major tasks in 3D point cloud processing—namely, 3D shape classification and semantic segmentation.

## 1 Introduction

The advancement of 3D point cloud acquisition techniques combined with the accessibility of acquisition devices has enabled the use of real-world 3D models in a variety of robotic applications, including autonomous driving, augmented reality, and robotics. 3D scanners, Light Detection and Ranging (LiDAR), RGB-D cameras (such as Kinect, RealSense, and Apple depth cameras) [1], and photogrammetry technologies allow the creation of very large and precise point clouds. Point clouds are the raw output of most 3D data gathering devices and serve as a versatile geometric representation of 3D data [2].

Effective point cloud analysis techniques are therefore essential to understanding of 3D targets. While hand-crafted features on point clouds have long been discussed in graphics and vision, the recent overwhelming success of convolutional neural networks (CNNs) for image analysis suggests that extending CNN insights to the point cloud domain could be beneficial. However, 3D point clouds are collections embedded in continuous space, unlike images, which are structured on regular pixel grids. This makes effective feature aggregation and message carrying methods among points in the cloud difficult to design, preventing the use of traditional deep network employed in computer vision. To form latent space mappings between input point coordinates and ground truth labels, pioneering research [3, 4] and subsequent work [5–11] have developed specialty modules for feature aggregation and message passing and yielded suitable neural networks for point cloud data.

The focus of this review is on the analysis of deep learning approaches for processing 3D point clouds for shape classification and semantic segmentation. We will also discuss some of the most prominent publicly available datesets used to handle diverse point cloud processing challenges. These

✉ Sushmita Sarker
sushmitasarkers@unr.edu

Prithul Sarker
prithuls@unr.edu

Gunner Stone
gunnerstone@unr.edu

Ryan Gorman
ryangorman@unr.edu

Alireza Tavakkoli
tavakkol@unr.edu

George Bebis
bebis@unr.edu

Javad Sattarvand
jsattarvand@unr.edu

[1] Department of Computer Science and Engineering, University of Nevada, Reno, NV, USA

[2] Department of Mining and Metallurgical Engineering, University of Nevada, Reno, NV, USA

datasets include ModelNet40 [12], ScanObjectNN [13], ShapeNet [14], S3DIS [15], Intra [16], Semantic3D [17], SemanticPOSS [18], and SydneyUrbanObjects [19]. Although there exists previous surveys of deep learning for 3D data, such as [20–25], this review specifically aims to bridge the gap by addressing techniques that previous surveys haven't comprehensively covered. it offers an immersive exploration of the latest frontiers in point cloud analysis. The primary objective of this paper is to equip readers with an extensive understanding of the diverse representations present in point clouds, with a particular emphasis on recent advancements within the field of raw point-based methodologies which have surged to the forefront of innovation. The key contributions of our paper are as follows:

1. Deep learning models for shape classification and semantic segmentation of 3D point clouds, covering the most up-to-date (2015–2023) advancements in this domain.
2. Our review goes beyond existing papers by encompassing all existing methods for point cloud classification and segmentation that have not been extensively discussed before.
3. We present a comprehensive taxonomy that encompasses both supervised and unsupervised approaches, including previously overlooked mesh-based methods. Our paper addresses the notable gaps in existing review papers by incorporating these previously unexplored methods.
4. Our analysis classifies and briefly discusses the myriad models available, each leveraging distinct representations and methodologies. This enables readers to grasp the diverse range of approaches within the context of their specific strengths and applications.
5. We conduct comprehensive comparisons of existing methods using multiple publicly available datasets and thoroughly expound upon the inherent strengths and limitations embedded within these diverse approaches.
6. Our paper includes a thorough discussion of the current challenges in the field and offers insightful directions for future research.

Our paper's novelty is evident not only in its coverage of recent advancements but also in its meticulous attention to previously overlooked areas in the literature. Additionally, the unique structure of our paper serves as a remarkable resource, catering to readers of all backgrounds—from newcomers seeking an approachable entry point to experts seeking a comprehensive taxonomy and insights into the latest deep learning methods for point cloud processing.

The structure of this paper is as follows: Sect. 2 introduces the datasets and evaluation metrics for the respective tasks. Moving forward, Sects. 3 and 4 reviews the state-of-the-art methods for 3D shape classification, while Sects. 5 and 6 provide comprehensive insights into the cutting-edge methods for semantic segmentation. Section 7 contains a quantitative assessment of several indicators as well as future research directions in this field and Sect. 8 concludes the paper.

## 2 Datasets and evaluation metrics

### 2.1 Datasets

A high quality dataset is crucial for both training and evaluating the effectiveness of machine learning algorithms. With the rise of deep neural networks, a reliable, well annotated and large datasets is even more crucial. In contrast to feature engineering used in traditional machine learning, deep network models rely on data and its annotations to extract appropriate feature embeddings. The purpose of 3D shape classification is to identify objects in a 3D point cloud [33–36] and assign a label to each discrete point. Thus, a large amount of well annotated training data is necessary for the model to train effectively.

In this paper, we collected a significant number of datasets to examine the performance of the state-of-the-art deep learning methods for various point cloud applications. Tables 1 and 2 lists some of the most common large-scale datasets currently used for 3D point cloud shape classification and segmentation.

The purpose of 3D shape classification is to identify objects in a 3D point cloud [33–36] and assign a label to each discrete point. Thus, a large amount of well annotated training data is necessary for the model to train effectively. For each dataset in Table 1, we present the establishment year, number of samples, number of classes and a brief description. We also categorize these datasets into two types: real-world datasets [13, 30] and synthetic datasets [12, 14]. Objects in the synthetic datasets exhibit no occlusion and are complete. Objects in real-world datasets may be partially occluded while background noise, outliers and point perturbations may be present in the data. ModelNet10 and ModelNet40 [12] are the most popular datasets employed for point cloud shape classification.

Table 2 provides an overview of commonly used large-scale datasets for 3D point cloud segmentation. These datasets are carefully curated and labeled to ensure representation of real-world scenarios and a wide range of object classes and scene types.

The datasets can be broadly classified into two groups: indoor datasets and outdoor datasets. In Table 2, we provide details such as the establishment year, number of points, number of classes, sensors used, and a brief description for each dataset. The data collection process involves various sensors, including RGB-D cameras [37], Mobile Laser Scanners (MLS) [27, 31, 38], Aerial Laser Scanners (ALS) [39, 40], and other 3D scanners [15]. Photogrammetry is often

**Table 1** Available point cloud datasets for classification

| Dataset | Year | Type | No. of Samples | No. of classes | Description |
|---|---|---|---|---|---|
| McGill 3D Shape Benchmark [26] | 2005 | Syn | 454 | 19 | Most of the models in this dataset were created using CAD modeling software and rest came from the Princeton Shape Benchmark |
| Sydney Urban Objects [19] | 2013 | RW | 588 | 14 | It contains scans generated by mobile platforms equipped with outdoor 3D scanner Velodyne LIDAR. Raw Velodyne range images are used for feature learning, and interpolated depth images are used for feature evaluation |
| Paris-rueMadame [27] | 2014 | RW | 642 | 26 | Data were obtained by the Mobile Laser Scanning (MLS) technology and correspond to a 160 m long street portion |
| ModelNet40 [12] | 2015 | Syn | 12311 | 40 | It is a complete and well-maintained collection of 3D CAD object models. The related point cloud data points are evenly sampled from the mesh surfaces, and they are further preprocessed by being moved to the origin and scaled into a unit sphere. It has 2468 meshes designated for testing and 9843 meshes for training |
| ModelNet10 [12] | 2015 | Syn | 4899 | 10 | A subset of ModelNet40 dataset which contains aligned orientation of the CAD models from 10 categories. The shapes are split into 80-20 ratios for training (3,991) and test (908) |
| ShapeNetCore [14] | 2015 | Syn | 51190 | 55 | ShapeNet, contains over 300 million models, with 220,000 classified into 3,135 classes. ShaperNetCore is a subset of the ShapeNet dataset |
| IQmulus [28] | 2015 | RW | – | 22 | This dataset contains 3D MLS data from a dense urban environment in Paris (France), composed of 300 million points |
| Object Scans [29] | 2016 | RW | 398 | 9 | This dataset includes more than 10,000 3D scans of real objects |
| ScanNet [30] | 2017 | RW | 12283 | 17 | ScanNet is an RGB-D video collection with 2.5M views across 1513 scenes annotated with mesh surfaces and 3D camera postures |
| Paris-Lille-3D [31] | 2017 | RW | 2479 | 50 | The dataset comprises of a point cloud created by the Mobile Laser System that was collected over around 2 kms in two French cities (Paris and Lille) |
| ScanObjectNN [13] | 2019 | RW | 2902 | 15 | ScanObjectNN is based on scanned indoor scene data. With 2902 distinct object instances, it has 15,000 objects that are divided into 15 categories. Due of the background, missing pieces, and deformations, it is a difficult point cloud classification dataset |

**Table 1** continued

| Dataset | Year | Type | No. of Samples | No. of classes | Description |
|---|---|---|---|---|---|
| Intra [16] | 2020 | Syn | 2025 | 2 | IntrA is an open-access 3D intracranial aneurysm dataset of 103 3D models of whole brain vessels and 1909 blood vessel segments, including 1694 healthy vessel segments and 215 aneurysm segments for diagnosis |
| ModelNet40-C [32] | 2022 | RW | 12308 | 40 | ModelNet40-C is built using ModelNet40 as a foundation. With 15 typical and realistic corruptions, it is the first comprehensive dataset for 3D point cloud corruption robustness |

*Syn* synthetic, *RW* real world type dataset employed to map the three-dimensional distance between objects.

## 2.2 Evaluation metrics

Many evaluation metrics have been proposed to evaluate different point cloud Application. To evaluate classification model usually the metric 'Accuracy' is used. In general, "accuracy" refers to the proportion of the model that predicts the correct outcome.

$$Accuracy = \frac{TP + TN}{TP + TN + FP + FN}$$

Here, TP, TN, FP, FN represent true positive, true negative, false positive and false negative respectively.

For 3D point cloud classification, overall accuracy (OA) and mean class accuracy (mAcc) are the most commonly used performance standards. "OA" evaluates the average accuracy across all test instances, while "mAcc" is used to evaluate the mean accuracy across all shape classes. Nowadays, dice coefficient (F1) score is also used as a criterion for performance evaluation in classification of 3D point clouds.

$$mAcc = \frac{1}{C} \sum_{c=1}^{C} Accuracy$$

$$F1 = \frac{2 \times precision \times recall}{precision + recall}$$

The F1 score is defined as the harmonic mean of precision and recall. Where,

$$Precision = \frac{TP}{TP + FP} \text{ and } Recall = \frac{TP}{TP + FN}$$

In 3D point cloud segmentation, several performance metrics are commonly used for evaluation, including mean intersection over union (mIoU), overall accuracy (OA), and mean class accuracy [15, 17, 31, 38]. These metrics provide insights into the quality of segmentation results. The IoU metric calculates the intersection over union between two sets, specifically the predicted bounding box (A) and the ground-truth bounding box (B). This overlap ratio is particularly relevant in segmentation tasks. The mIoU is the average IoU computed for each category, providing an overall measure of segmentation accuracy. The IoU can be computed using the following equation:

$$IoU\ Score(A, B) = \frac{|A \cap B|}{|A \cup B|}$$

These metrics enable a comprehensive assessment of the accuracy and effectiveness of 3D point cloud application algorithms.

## 3 3D point cloud classification

The subject of a 3D point cloud shape classification is to produce a label for the entire point cloud determining the shape of the object it contains. Analogous to 2D image classification, methods for 3D shape classification tasks usually follow two main stages. First the embedding of each point is learned in order to generate a global embedding with an aggregation encoder. Next the embedding is passed through several fully connected layers to obtain the final shape label.

Based on the input data type, point cloud classification models can be generally divided into five major classes, i.e., mesh based methods, projection-based methods, volumetric-based methods, hybrid methods, and raw point-based methods. Mesh data is a common method for representing 3D shapes in computer graphics, consisting of interconnected vertices, edges, and faces. While mesh data provides an efficient way to store and render 3D models, it also presents

**Table 2** Available point cloud datasets for segmentation

| Name | Year | #Points | Classes | Sensors | Description |
|---|---|---|---|---|---|
| Oakland [41] | 2009 | 1.61M | 5 | MLS | The data set consists of two subsets (part2 and part3), each having a unique local reference frame and 100,000 3-D points in each file. Filtered, labeled, and remapped from 44 into 5 labels, the training/validation and testing data have 36932/91579 and 1.3 M points, respectively |
| IQmulus [28] | 2013 | 300M | 22 | MLS | The database comprises 300 million points of 3D MLS data from a dense metropolitan setting in Paris, France |
| Paris-rue-Madame [27] | 2014 | 20M | 27 | MLS | The Paris-rue-Madame dataset contains 3D Mobile Laser Scanning (MLS) data from rue Madame, a street in the 6th Parisian district (France) which comprises an approximately 160 m long street section |
| ScanNet [30] | 2017 | – | 20 | RGB-D | ScanNet is a dataset of 2.5 million RGB-D images of 1513 scans collected in 707 different places. It contains richly annotated RGB-D scans of real-world surroundings |
| S3DIS [15] | 2017 | 695.9M | 13 | Matterport | The dataset is collected in 6 large-scale indoor areas and covers over 6,000 square meter. It has over 70,000 RGB images. It contains 272 3D room scenes of 13 categories |
| Semantic3D [17] | 2017 | 4B | 8 | MLS | Semantic3D is a point cloud database of scanned urban outdoor scenes with over 3 billion points. It has 15 training and 15 test scenes with 8 class labels annotated. This extensive set of 3D point clouds with labels includes a variety of urban scenes |
| Paris-Lille-3D [31] | 2018 | 143.1M | 50 | MLS | Paris-Lille-3D is a metropolitan 3D point clouds dataset with 140 million points spanning a distance of 2 kms in two separate cities. The items were manually segmented, and each one was assigned to one of 50 classes |
| DublinCity [42] | 2019 | 260M | 13 | ALS | The dataset contains about 260 million labeled laser scanning points out of 1.4 billion points. These are carefully annotated into approximately 100,000 items from the Dublin LiDAR point cloud in 2015 |
| SemanticKITTI [38] | 2019 | 4549M | 28 | MLS | SemanticKITTI is a large-scale outdoor-scene dataset based on the KITTI Vision Benchmark for point cloud semantic segmentation. It contains 43552 scans of outdoor scenes of 28 classes, of which 23201 are used for training, while the rest of 20351 are reserved for testing |
| PreSIL [43] | 2019 | 3135M | 24 | LiDAR | It has more than 50,000 frames and includes high-definition images with full resolution depth information, semantic image segmentation, point-wise segmentation, and in-depth annotations for every vehicle and person |
| SensatUrban [44] | 2020 | 2847M | 13 | UAV Photogrammetry | The collection includes sizable portions of two UK cities, covering around 6 square kilometers of the city landscape. Each 3D point in the dataset is assigned to one of 13 semantic classifications |

**Table 2** continued

| Name | Year | #Points | Classes | Sensors | Description |
|---|---|---|---|---|---|
| Swiss3DCities [45] | 2020 | 226M | 5 | UAV Photogrammetry | Swiss3DCities is a new outdoor urban 3D pointcloud dataset, covering a total area of 2.7 square kilometers, and is sampled from three Swiss cities with different characteristics. It is manually annotated for semantic segmentation using per-point labels |
| LASDU [46] | 2020 | 3.12M | 5 | ALS | An enhanced large-scale geometric dataset based on ShapeNetCore with accurate semantic region annotations, and detailed per-point labeling. It includes 31963 models in 16 different shape categories |
| Campus3D [47] | 2020 | 937.1M | 24 | UAV Photogrammetry | It is a well annotated 3D point cloud dataset for different outdoor scene comprehension tasks. The dataset was created by photogrammetric processing of unmanned aerial vehicle (UAV) images taken at the National University of Singapore (NUS) |
| SemanticPOSS [18] | 2020 | 216M | 14 | MLS | A huge number of dynamic instances are included in 2988 different and challenging LiDAR scans that make up the SemanticPOSS dataset for 3D point cloud segmentation. It employs the same data structure as SemanticKITTI and is acquired at Peking University |
| Toronto-3D [48] | 2020 | 78.3M | 8 | MLS | Toronto-3D is a large-scale urban outdoor point cloud dataset that was collected for semantic segmentation by an MLS system in Toronto, Canada. This dataset has 78.3 million points and spans a distance of nearly 1 km of road |
| DALES [40] | 2020 | 505M | 8 | ALS | The Dayton Annotated LiDAR Earth Scan (DALES) data set is a new, extensive aerial LiDAR data set with more than 500 million hand-labeled points covering an area of 10 square kilometers and eight object categories |
| RELLIs-3D [49] | 2021 | 176.1M | 20 | LiDAR | RELLIS-3D is a multi-modal dataset for off-road robotics. It was gathered in an off-road setting and includes 6235 photos and 13,556 LiDAR scan annotations |
| WADS [50] | 2021 | 3.6B | 22 | MLS | It is the first multi-modal dataset with dense point-wise labeled sequential LiDAR scans acquired in harsh winter conditions |
| H3D [51] | 2021 | 73.4M | – | ALS | The H3D (Honda Research Institute 3D) dataset is a large-scale RGB-D dataset which was released by the Honda Research Institute and contains over 100,000 images of 244 objects in various cluttered scenes, with 6D pose annotations for each object instance |
| SynLIDAR [52] | 2022 | 19,482M | 32 | Synthetic LiDAR | SynLiDAR is a 19 billion point synthetic LiDAR sequential point cloud dataset with point-by-point annotations of 32 semantic classes |
| STPLS3D-Real [53] | 2022 | – | 6 | UAV Photogrammetry | The dataset includes outdoor images taken at four actual locations. The aerial images were taken using a crosshatch-style flight pattern, with specified overlaps of 75–85% and flight heights of 25–70 ms |
| STPLS3D-SyntheticV1 [53] | 2022 | – | 7 | UAV Photogrammetry | For semantic and instance segmentation applications, STPLS3D is an extensively annotated synthetic 3D aerial photogrammetry point cloud dataset. It contains 16 square kilometers of landscape and up to 18 fine-grained semantic categories |

*ALS* Airborne Laser Scanning; *MLS* Mobile Laser Scanning

Table 3 Comparative 3D point cloud classification result on various available datasets for projection-based, volumetric-based, mesh-based and Hybrid methods

| Model name | Year | Rep. | ModelNet 40 (OA) | (mAcc) | ModelNet 10 (OA) | (mAcc) | ScanObjectNN (OA) | (mAcc) | SydneyUrbanObjects (F1) |
|---|---|---|---|---|---|---|---|---|---|
| **Mesh-based Methods** | | | | | | | | | |
| Geometry Image [84] | 2016 | PM | 83.90% | 51.30*% | 88.40% | 74.90*% | – | – | – |
| Cross-atlas [85] | 2019 | PM | 87.50% | – | 91.20% | – | – | – | – |
| SNGC [86] | 2019 | PM | 91.60% | – | – | – | – | – | – |
| MeshNet [54] | 2019 | PM | 91.90% | 81.90*% | – | – | – | – | – |
| MeshWalker [55] | 2020 | PM | 92.30% | – | – | – | – | – | – |
| CurveNet [57] | 2020 | CM + Vol. | 90.70% | – | 94.20% | – | – | – | 79.30% |
| PolyNet (unsqueezed) [56] | 2021 | PM | 92.42% | 82.86*% | 94.93% | 84.62*% | – | – | – |
| RepSurf-U [58] | 2022 | TCM | 94.70% | 91.70% | – | – | 84.60% | 81.90% | – |
| **Projection-based Methods** | | | | | | | | | |
| MVCNN [60] | 2015 | 80 views | 90.10% | – | – | – | – | – | – |
| MHBN [62] | 2018 | 6 views | 93.10% | 94.70% | 95.00% | 95.00% | – | – | – |
| GVCNN [61] | 2018 | 8 views | 93.10% | 84.50% | – | – | – | – | – |
| CNN + LSTM + voting [66] | 2019 | 8 views | 91.05% | – | 95.29% | – | – | – | 75.30% |
| Dominant Set clustering [65] | 2019 | 24 views | 93.80% | – | – | – | – | – | – |
| SimpleView [87] | 2021 | 6 view | 93.90% | 91.80% | – | – | 80.50% | 75.50% | – |
| MVTN [67] | 2021 | 20 views | 93.50% | 92.20% | – | – | 82.80% | – | – |
| MVACPN [68] | 2021 | 6 views | 93.64% | 91.53% | – | – | – | – | – |
| **Volumetric-based Methods** | | | | | | | | | |
| 3D ShapeNet [12] | 2015 | Vol. | – | 77.32% | – | 83.54% | – | – | – |
| VoxNet [75] | 2015 | Vol. | – | 83.00% | – | 92.00% | – | – | 73.00% |
| FPNN [88] | 2016 | Vol. | 88.40% | – | – | – | – | – | – |
| VRN Ensemble [73] | 2016 | Vol. | – | 95.54% | – | 97.14% | – | – | – |
| Binary VoxNetPlus [89] | 2017 | Vol. | – | 85.47% | – | 92.32% | – | – | 75.50% |
| LightNet [90] | 2017 | Vol. | – | 86.90% | – | 93.39% | – | – | 76.00% |
| LP-3DCNN [91] | 2019 | Vol. | – | 92.10% | – | 94.40% | – | – | – |
| MRCNN [71] | 2019 | Vol. | – | 86.20% | – | 91.30% | – | – | – |
| MS-VDNN [92] | 2020 | Vol. | – | 92.93% | – | 95.30% | – | – | – |
| AF2M [72] | 2021 | Vol. | 93.16% | – | – | – | – | – | – |
| **Hybrid Methods** | | | | | | | | | |
| FusionNet [93] | 2016 | Vol. + 60 views | – | 90.80% | – | 93.11% | – | – | – |
| PVNet [77] | 2018 | PC + 12 Views | 93.20% | – | – | – | – | – | – |

**Table 3** continued

| Model name | Year | Rep. | ModelNet 40 (OA) | (mAcc) | ModelNet 10 (OA) | (mAcc) | ScanObjectNN (OA) | (mAcc) | SydneyUrbanObjects (F1) |
|---|---|---|---|---|---|---|---|---|---|
| 3DmFV-Net [76] | 2018 | PC + Vol. | 91.60% | – | 95.20% | – | – | – | 76.00% |
| PointGrid [74] | 2018 | PC + Vol. | 92.00% | 88.90% | – | – | – | – | – |
| PVRNet [78] | 2019 | PC + 12 Views | 93.60% | – | – | – | – | – | – |
| PointView-GCN [80] | 2021 | PC + 20 views | <u>95.40%</u> | – | – | – | 85.50% | – | – |
| DSPoint [79] | 2022 | PC + Vol. | 93.50% | – | – | – | – | – | – |
| Point-Voxel Transformer [81] | 2022 | PC + Vol. | 94.00% | – | – | – | – | – | – |
| GSV-Net [94] | 2022 | PC + 3 Views | 92.70% | – | – | – | – | – | – |
| PointCMT [82] | 2022 | PC + Image | 94.40% | 91.20% | – | – | **86.40%** | **84.40%** | – |
| EQ-PointNet++ (SSG) [83] | 2022 | PC + Vol. | 93.20% | – | – | – | – | – | – |

*OA* Overall Accuracy; *mAcc* Mean Accuracy

Results with '*' sign show mean average precision score. The symbol '–' is used to indicate that results are unavailable. The methods are arranged in chronological order within their corresponding categories. The top-performing methods in each category have been highlighted in bold, while the method(s) achieving the best overall performance across all categories are underlined

---

challenges due to its inherent complexity and irregularity. Projection-based methods project the unstructured point cloud into 2D images (rasterization) to extract features. These features are then fed into 2D or 3D convolutional networks. In volumetric-based methods, the point cloud is discretized into a regular grid, creating a volumetric representation of the data. In contrast, point-based methods directly work on raw points in the cloud. Point-based methods have become increasingly popular since they reduce the computational complexity of the network without any explicit information loss. Papers that combine the benefits of both point and projection methods are referred to in this review as "hybrid" techniques.

In Sect. 3, we have discussed the various models for 3D shape classification with a focus on input data representation. Meanwhile, Sect. 4 presents an in-depth analysis of methodologies exclusively reliant on raw point clouds as their input. Table 3 provides a comprehensive comparison of the aforementioned methods for 3D point cloud classification across various datasets. The methods have been categorized based on the representation of the input point cloud utilized in each approach. Furthermore, the methods are arranged in chronological order within their respective categories. The evaluation of each method's performance is based on metrics such as overall accuracy (OA), mean accuracy (mAcc), mean average precision (mAP), and F1 score.

### 3.1 Mesh-based methods

A mesh is the most widely used structure for representing surfaces in computer graphics and is comprised of a set of faces and vertices that define surfaces on a three-dimensional shape. As a result, this representation carries structural information about the object's surface. Furthermore, by pruning vertices in a mesh and removing extraneous data, mesh-based representations provide a memory-efficient way to store complete geometry details. However, this representation is often overlooked as a suitable input modality for deep learning algorithms. This could be attributed to the fact that a 3D mesh does not provide a grid-like pattern for representing the data to be used in CNNs. In addition, weight sharing in mesh-based approaches presents a difficult challenge because of changes in the number of vertices, the permutation of adjacent vertices, and their pairwise distances.

To learn 3D shape representation from mesh data, Feng et al. introduced a mesh neural network as MeshNet in [54]. This approach introduces face-unit and feature splitting and proposes a general architecture with usable and efficient building blocks. The face unit takes as input the features of the vertices and edges that make up a single face and applies convolutional operations to learn representations of that face. These representations are then combined to form higher-level representations of the mesh. MeshNet can effectively manage mesh irregularity and complexity concerns for 3D shape representation. Alternatively, MeshWalker [55] directly learns the shape from a given mesh without the need for changing the mesh data representations. This is done by examining the geometry and topology of the mesh by randomly traversing its surface via a number of walks. Each walk is arranged as a list of vertices and imposes some degree of regularity on the mesh.

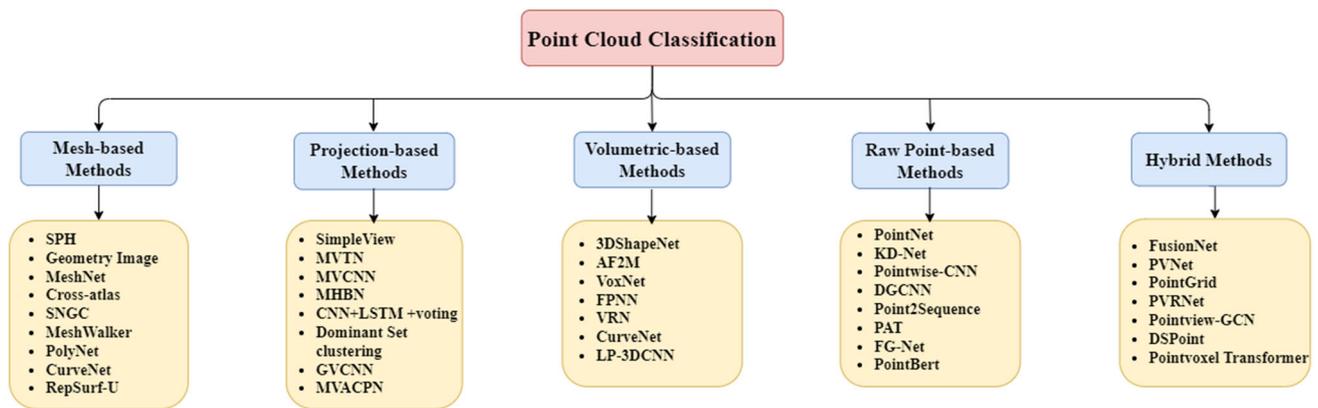

**Fig. 1** A taxonomy of deep learning methods for 3D point cloud classification

Instead of randomly processing vertices, PolyNet [56] can efficiently learn and extract features from a polygon mesh representation of 3D objects using continuous polynomial convolution (PolyConv). A PolyConv is a polynomial function with learnable coefficients that develops continuous distributions about the features of vertices sharing the same ploygonal face. These convolutional filters distribute appropriate weights among the vertices in the local patches formed by each vertice and its neighboring vertices on the surface of the polygon. This process is invariant with respect to the quantity of neighboring vertices, their permutations, the pairwise distances between them and the choice of central vertex in a local patch. Although mesh-based learning has gained significant popularity in the field of computer vision, it still poses some challenges. This can be attributed to the fact that 3D meshes do not conform to the grid-like structure that is typically used for representing data in convolutional neural networks (CNNs).

In order to address the challenges with meshes, recent advancements in this area have drawn inspiration from triangle meshes and curvature maps used in computer graphics. Muzahid et al. [57] recently introduced a new approach, called CurveNet, using curvature directions to capture geometric features from polygon meshes as inputs to a 3D CNN. The data structure of CurveNet enables object class label prediction by learning perceptually significant and salient features. Curvature directions provide detailed surface information from 3D objects, allowing the model to generate more precise and discriminative features for accurate object recognition. Similarly, Ran et al. [58] proposed a novel approach, Triangular RepSurf, that draws inspiration from triangle meshes in computer graphics. It can be computed by predefined geometric priors after surface reconstruction. RepSurf has two variants: Triangular RepSurf and Umbrella RepSurf. Triangular RepSurf represents each local region as a triangle mesh, while Umbrella RepSurf represents each local region as an umbrella-shaped structure [59].

Table 3 shows that RepSurf-U [58] outperforms all other mesh-based methods on ScanObjectNN [13] dataset, achieving an OA of 84.6% and an mAcc of 81.9%. PolyNet (unsqueezed) [56] achieves the highest OA and mAcc scores on ModelNet 40 [12] dataset with 92.4% and 82.86% respectively. On the other hand, among all other methods, CurveNet [57] achieves the highest F1 score of 79.3% on SydneyUrbanObjects dataset.

### 3.2 Projection-based methods

Projection-based methods are popular approaches for analyzing unstructured 3D point clouds. By projecting the point cloud onto multiple two-dimensional (2D) planes (or views), the data can be more effectively analyzed using standard image processing techniques. The resulting view-wise features can be collected and concatenated to produce a more precise classification of the point cloud's shape. However, one significant challenge faced by projection-based methods is in combining multiple view-wise features into a distinct global representation that accurately captures the overall structure of the point cloud.

Su et al. introduced a novel approach for processing point clouds called Multi-View Convolutional Neural Network (MVCNN) [60]. This approach involves representing point clouds as a collection of 2D images captured from multiple views obtained at different angles. Features are extracted from these different views and then combined into a global descriptor through max-pooling. However, one potential drawback of max-pooling is that it only retains the most important parts of each view, resulting in a loss of information. While MVCNN does not explicitly differentiate between various views, it is beneficial to have some understanding of the relationships between them.

One such method that specifically attempts to etablish relationships between views of a point cloud is Group-view Convolutional Neural Network (GVCNN) [61]. GVCNN splits the set of views into groups based on their discrimination scores, thereby leveraging the relationship between them for better results. Other techniques such as Multi-View Harmonized Bilinear Pooling Network (MHBN) [62] use harmonized bilinear pooling to combine local convolutional features into a more compact and discriminative global descriptor. Meanwhile, Yang et al. proposed a method that exploits the inter-relationships between views and regions using a relation network to generate a more accurate 3D object representation [63]. Unlike earlier techniques, Wei et al. presented the View Graph Convolutional Network (View-GCN) which employs a directed graph to consider many views simultaneously [64]. Other strategies such as Dominant Set Clustering Network (DSCN) [65], and Learning Multi-View 3D Object Recognition (LMV3D) [66] have also been proposed to improve recognition accuracy.

Despite the popularity of methods that utilize raw point data such as PointNet [3], some projection-based methods have yielded promising classification results. For instance, Abdullah et al. recently introduced a differentiable module that predicts the optimal viewpoint for a multi-view network [67]. MVTN overcomes the static nature of existing projection-based techniques by utilizing adaptive viewpoints, which it learns to regress. These viewpoints are rendered with a differentiable module to train the task-specific network end-to-end. This results in the most appropriate views for the task at hand.

In another study, Wang et al. [68] presented a multi-view attention-convolution pooling network (MVACPN) framework using Res2Net [69] to extract features from several 2D views. MVACPN effectively resolves the issues of feature information loss caused by feature representation and detail information loss during dimensionality reduction by employing the attention-convolution pooling method.

The results in Table 3 demonstrate that the MHBN [62] method outperformed other projection-based methods with the highest mean accuracy on the ModelNet 40 [12] and ModelNet 10 datasets, while [66] obtained the highest overall accuracy on ModelNet10 among all other approaches.

### 3.3 Volumetric-based methods

Another approach to produce structured data for processing in traditional CNN architectures is to convert the point cloud into a regular 3D grid of cubic voxels using a process called voxelization. Each point in the point cloud is assigned to the closest voxel center in the 3D grid, resulting in a volumetric representation of the point cloud.

Wu et al. [12] introduced 3D ShapeNet, a deep belief-based convolutional network that learns the distribution of points from diverse 3D shapes. The network represents the points as a probability distribution of binary variables on voxel grids. Despite promising results, these methods struggle to scale to dense 3D data, as memory consumption increases cubically with resolution.

To enable hierarchical learning of features, Ghadai et al. in [71] presented a flexible multi-level unstructured voxel representation of spatial data in their MRCNN framework. This method uses a multi-level voxelization framework, described as a binary occupancy grid at two levels. This is done to represent a 3D object with two distinct user-specified resolutions of voxel grids. Despite the lack of structure in the multi-level data representation, MRCNN can successfully learn features, allowing for more effective and efficient 3D shape classification.

To exclusively take voxelized data as input in an end-to-end encoder-decoder CNN architecture, Cheng et al. suggested a similar technique called (AF)2-S3Net [72]. This method uses a multi-branch attentive feature fusion module to learn both global contexts and local features in the encoder. To promote generalizability, an adaptive feature selection module with feature map re-weighting is utilized on the decoder side to actively emphasize contextual information from a feature fusion module.

Table 3 illustrates that AF2M [72] achieved the highest overall accuracy on the ModelNet40 dataset. However, they did not report their mAcc result. On the other hand, VRN Ensemble [73] only provided their mAcc results for both ModelNet40 and ModelNet10 [12], which are highlighted in bold and underlined in the table as they represent the highest results for those datasets among all other approaches. Nevertheless, it is worth noting that none of the papers recorded their findings for the ScanObjectNN dataset.

### 3.4 Raw point-based methods

To address information loss and maintain point cloud details, Raw Point-based Methods offer a promising alternative in point cloud processing. These methods operate directly on the raw point cloud data, avoiding the need for transformation into other representations. PointNet [3] pioneered this approach by consuming unordered point sets and achieving permutation invariance through symmetric functions. This novel approach has facilitated the accurate analysis of raw point cloud data, eliminating the need for conversion into other representations. Raw point processing has been a major focus of recent models. Since a substantial body of work exists on raw point processing, we will explore the different learning strategies employed in these methods in Sect. 4.

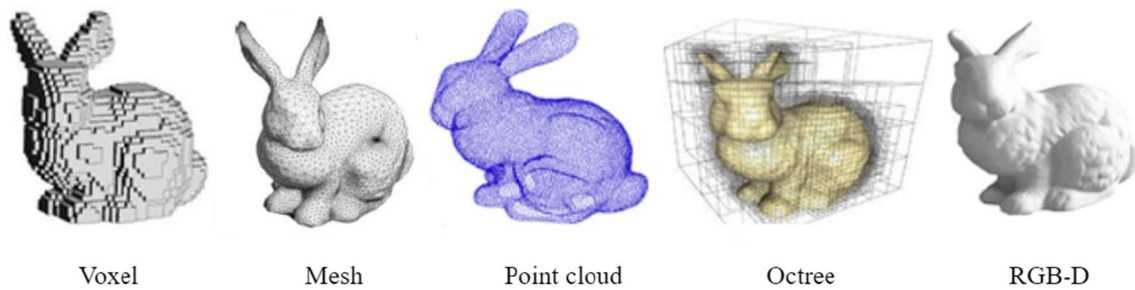

**Fig. 2** The Stanford Bunny [70] model in different three-dimensional representations

## 3.5 Hybrid methods

As discussed earlier, point cloud classification techniques fall under three broad categories of projection-based, voxel-based or point-based neural network(NN) models to handle 3D input. All these approaches, however, are computationally inefficient. The memory usage and computation cost of voxel-based models expand cubically with input resolution. In point-based networks, the majority of the computational cost is on processing the sparse input points to produce data conducive for the remainder of the network. This process often leads to poor memory localization, rather than effective feature extraction. Approaches that combine a variety of input data modalities are known as Hybrid methods. These methods are relatively new, and an increasing number of researchers are investigating various challenging questions in this domain.

Integrating voxel-based learning with point-based learning into a unified framework has been the subject of recent developments in point cloud classification. For example, PointGrid, presented in [74], is a hybrid network that combines the point and grid representations. To retrieve the specifics of local geometry, it uses a 3D CNN to learn the grid cells with fixed points. The PointGrid model employs the same transformation mechanism as VoxNet [75], but it can better describe scale changes, minimizes data loss, and takes up less memory.

The development of real-time algorithms for 3D point cloud classification is highly challenging due to the large size and complexity of point clouds. To overcome this challenge, Ben-Shabat et al. [76] utilized the 3D modified Fisher Vector (3DmFV) approach to convert an input point cloud into 3D grids, followed by CNNs to extract features and fully connected layer to classify them in real-time. The 3DMFV method is an extension of the widely-used 2D modified Fisher Vector (mFV) method for image classification.

Specifically, it extends the mFV method to 3D point clouds by encoding the gradient information of the 3D grids using a Gaussian mixture model (GMM) and computing the Fisher vector (FV) representation of the point cloud. In contrast to the 3DmFV approach, PVNet [77] uses an embedding network to project high-level global features collected from multi-view images into the subspace of point clouds, which are then blended with point cloud features using a soft attention mask. Finally, for fused features and multi-view features, a residual connection is used to achieve shape recognition. To further improve accuracy, You et al., proposed to leverage the relationship between a 3D point cloud and its multiple views through a relation score module in PVRNet [78].

In DSPoint [79], Zhang et al. introduced a dual-scale point cloud recognition approach that combines local features and global geometric architecture. Unlike conventional designs, DSPoint operates concurrently on voxels and points, extracting local and global features. The network disentangles point features through channel dimensions, enabling dual-scale processing. It utilizes pointwise convolution for fine-grained geometry parsing and voxel-wise global attention for long-range structural exploration. To align and blend the local–global modalities, a co-attention fusion module is designed for feature alignment, facilitating inter-scale cross-modality interaction by incorporating high-frequency coordinate information. In contrast, PointView-GCN [80] uses multi-level Graph Convolutional Networks (GCNs) to hierarchically aggregate shape features from single-view point clouds. It captures geometrical cues and multiview relations for 3D shape classification by leveraging partial point cloud data from multiple views.

Voxel-based models have regular data locality and can efficiently encode coarse-grained features. On the other hand point-based networks preserve the accuracy of location information with the flexible fields. Inspired by this, Zhang et al. in [81] proposed a hybrid point cloud learning architecture, called PointVoxel Transformer. The authors used the Sparse Window Attention (SWA) module to gather coarse-grained local features from nonempty voxels. The module not only bypasses the expensive irregular data structuring and invalid empty voxel computation, but also obtains linear computational complexity with respect to voxel resolution. In another recent work by Yan et al., called PointCMT [82], both image and point cloud data is used to train the model for shape analysis tasks. This approach combines the advantages of both modalities, leveraging the rich texture information from images and the geometric structure from point clouds.

In [83], the authors introduced the Embedding-Querying (EQ-Paradigm), a unified approach for 3D point cloud understanding. EQ-Paradigm combines different task heads with existing 3D backbone architectures, offering advantages such as a unified framework for tasks like object detection, semantic segmentation, and shape classification. It seamlessly integrates with diverse 3D backbone architectures and efficiently handles large point clouds.

Table 3 Shows the qualitative evaluation of classification results of hybrid methods on various publicly available datasets. The best-performing method on ModelNet 40 dataset was PointView-GCN, which achieved an OA of 95.40%, while PointCMT achieved the highest mAcc and F1 scores among other methods on ScanObjectNN dataset.

## 4 Learning strategies for point based methods in classification

Since the development of PointNet, numerous models have emerged that can process raw point data directly without information loss [3, 95–97]. These models employ diverse techniques and network architectures to handle unstructured data. In this section, we will discuss about the learning strategies that these models have adopted for processing raw points. We provide a detailed discussion of each category and highlight their key differences and commonalities. Generally, methods in this category can be broadly categorized into two groups depending on the type of supervision used during training. Figure 3 provides a comprehensive categorization of raw point-based approaches for point cloud shape classification.

**Supervision in training** Supervision in point cloud processing involves training neural networks with labeled point clouds to make predictions on unlabeled ones. It can be divided into two categories: supervised and unsupervised training. Supervised methods use labeled data to teach the model how to predict outputs for new point clouds. Unsupervised methods, on the other hand, identify patterns and structures in the input data without prior knowledge of the output. Both supervised and unsupervised methods are crucial in point cloud processing, depending on the availability and quality of labeled data. More details about supervised and unsupervised methods of raw point cloud processing are discussed in Sects. 4.1 and 4.2 respectively.

Table 4 presents a comprehensive comparison of raw point-based methods for 3D point cloud classification across various datasets. The methods are organized chronologically within their respective categories. The table includes the number of parameters reported by each paper and specifies whether the model used only the point cloud or also incorporated point features such as normals as input. The evaluation of each method's performance is based on metrics such as overall accuracy (OA), mean accuracy (mAcc), and F1 score. Notably, the results for the Intra dataset are showcased and obtained from [98].

### 4.1 Supervised training

Supervised learning for point cloud is a powerful approach for processing and analyzing 3D data. It involves training the system on labeled point clouds to extract meaningful information such as object classification, semantic segmentation, and registration [3, 99]. The algorithm is continuously improved as it compares its predictions with the desired output, allowing for more accurate results with each iteration. However, supervised learning requires large amounts of labeled data, which can be costly to obtain.

Supervised learning approaches can be classified into seven categories: pointwise MLP, hierarchical-based, convolution-based, RNN-based, graph-based, transformer-based, and other methods. These categories can be further grouped into feedforward and sequential training based on the model architecture and data processing method.

**Feed-forward training** It is an extensively used technique for processing point clouds, where the individual points of the point cloud are passed through multiple layers in a neural network to generate activation maps for successive layers. This allows the model to capture complex relationships by transforming the data through non-linear transformations. Based on the operations performed on points in each layer, this group encompasses the following methods: multilayer perceptron (MLP)-based, convolution-based, hierarchical-based, and graph-based architectures.

**Sequential training** It is a type of training method, the model is trained on a sequence of input data. In point cloud processing, the input data is treated as ordered points or patches, processed in sequence. Unlike feed-forward training, where data flows from input to output, sequential training uses the output from one time step as the input for the next. This approach is beneficial in point cloud processing as it allows the model to process local patches and predict features for the next point. Sequential training is commonly used in recurrent neural networks (RNNs) and transformer-based architectures designed to process sequential data.

Supervised learning is a crucial element of point cloud processing pipelines, particularly in cases where high accuracy is essential. In the following sections, we will provide an in-depth discussion of the various network architectures that are utilized for feature learning of individual points, with supervised learning being the primary technique.

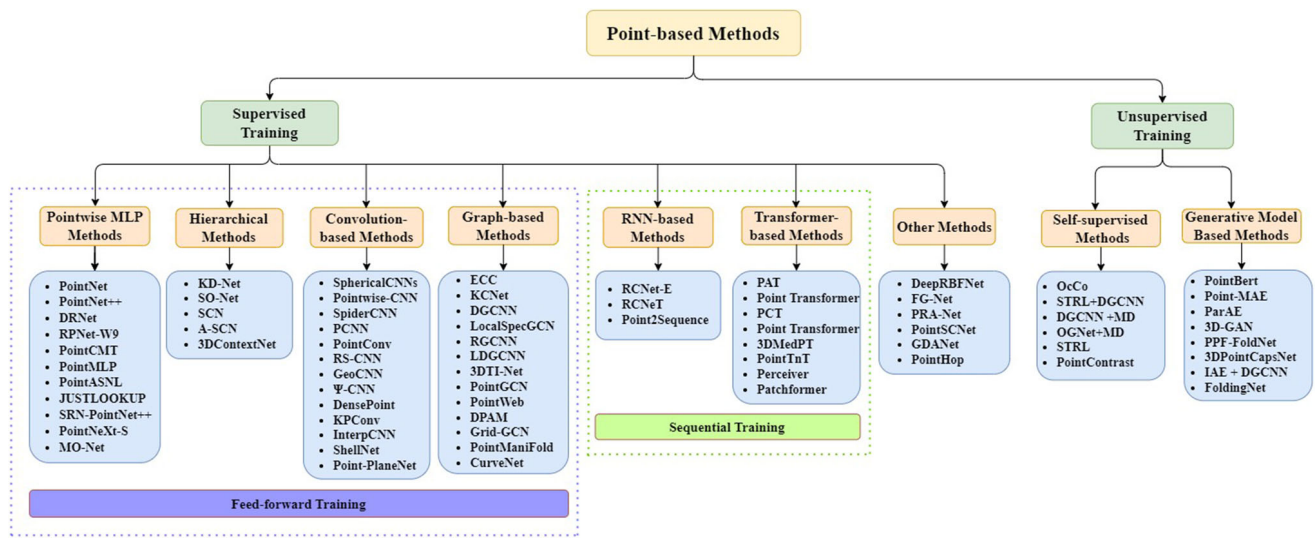

**Fig. 3** A taxonomy of deep learning approaches for raw point-based 3D point cloud classification

### 4.1.1 Multi-layer perceptron (MLP) methods

This method is based on fully connected layers that process each point independently. The network takes a point cloud and applies a set of transforms and shared MLPs to generate features. These features are then aggregated to yield a global representation using max-pooling that describes the original input cloud. Another MLP classifies that global representation to produce output scores for each class.

PointNet [3], in particular, uses multiple MLP layers to learn pointwise features independently and a max-pooling layer to extract global features. The local structural information between points cannot be captured since features are learned independently for each point in PointNet. As a result, Qi et al. presented PointNet++ [4], a hierarchical network that captures complex geometric patterns in the neighborhood of each point. PointNet++ is inspired by standard CNNs, which use a stack of convolutional layers to capture features at different scales. The points within a sphere centered at x are defined as the local region of the point x. In particular, one set abstraction level contains a sampling layer, a grouping layer to identify local regions, and a PointNet layer. PointNet [3] and PointNet++ [4] prompted a lot of follow-up work due to their easy implementation and promising performance. Mo-Net [100] has a similar design to PointNet, but it takes a fixed collection of moments as input. SRINet [101] uses a PointNet-based backbone to extract a global feature and graph-based aggregation to extract local features after projecting a point cloud to generate rotation invariant representations.

Yan et al. [7] used an Adaptive Sampling (AS) module in their work, PointASNL, to adaptively adjust the coordinates and attributes of points. They sampled these points using the Farthest Point Sampling (FPS) technique and presented a local-non-local (L-NL) module to capture the local and long-range relationships of the sampled points. Duan et al. [97] proposed utilizing MLP to learn structural relational properties between distinct local structures using a Structural Relational Network (SRN). Lin et al. [102] used a lookup table to speed up the inference process for both the input and function spaces learned by PointNet. On a consumer grade computer, the inference time for the ModelNet and ShapeNet datasets is 1.5 ms and 32 times faster than PointNet. In RPNet, Ran et al. [103] studied the capabilities of local relation operators and developed the group relation aggregator (GRA), a scalable and efficient module for learning from both low-level and high-level relations. The module calculates a group feature by aggregating the features of inner-group points that are weighted by geometric and semantic relations. RPNet contains approximately a third of the parameters of PointNet++ and double the computation speed.

Previous works have mainly focused on utilizing advanced local geometric extractors such as convolution, graphs, and other mechanisms to capture 3D geometries. However, these methods can lead to increased computational costs and memory usage. To address this challenge, Ma et al. [104] developed PointMLP, a pure residual MLP network that does not rely on "complex" local geometrical extractors. Despite this simplicity, PointMLP performs well due to highly optimized MLPs. They developed a lightweight local geometric affine module that adaptively modifies the point feature in a local region to boost efficiency and generalization ability. PointMLP trains two times faster and tests seven times faster than the current models. PointNext [105] overcomes the limitation of PointNet++ [4] by utilizing a thorough analysis of model training and scaling techniques. The authors add sep-arable MLPs and an inverted residual bottleneck design to PointNet++ to facilitate effective and efficient model scaling. In PointStack [106], the authors proposed a method that utilizes multi-resolution features and learnable pooling to extract meaningful features from point cloud data. The multi-resolution features capture the underlying structure of the point cloud data at different scales, while the learnable pooling enables the system to dynamically adjust the pooling operation based on the features.

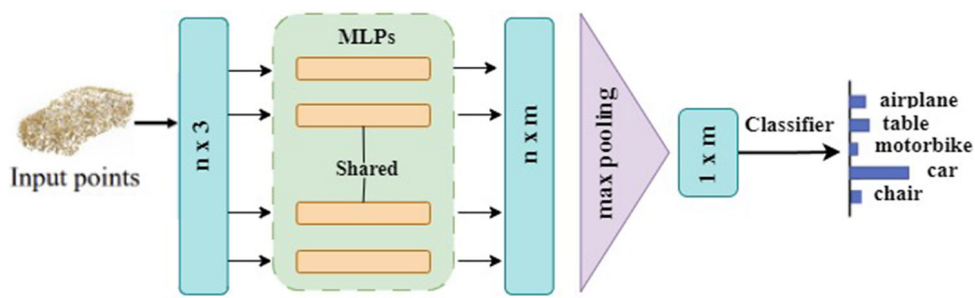

**Fig. 4** A simplified architecture of PointNet [3] where parameters n and m denote point number and feature dimension, respectively

Table 4 shows that PointStack achieves the best results for both the ModelNet 40 and ScanObjectNN datasets, while PointASNL achieves the best result for the ModelNet 10 dataset, and PointNet++ for the Intra dataset among all the MLP-based methods.

### 4.1.2 Convolutional methods

The architecture of convolution networks is an emulation of biological processes and is closely related to the organization of the visual cortex in animals. In this architecture, each cortical neuron primarily responds to inputs within its receptive field. Multiple neurons with overlapping receptive fields respond to the entire field at a particular location. To extract features from low-level to high-level features, convolutional networks are stacked with convolution layers, rectified linear units, and pooling layers. The strengths of convolutional networks include shared weights, translation invariance, and feature extraction, as demonstrated in several works, such as ApolloCar3D [107], and Semantic3D [17]. VoxNet [75] illustrated the use of 2D grid kernels for processing 3D point cloud data. However, due to the irregularity of point clouds, constructing convolution kernels for 3D point clouds presents greater challenges compared to 2D counterparts. Modern 3D convolution methods can be categorized as discrete or continuous based on the nature of the convolution kernel used.

**Discrete convolution** Discrete convolution for point cloud processing involves defining a convolutional kernel on a regular grid based on a set of surrounding points that are located within a certain radius from the center point. This technique leverages the structural properties of point clouds, which can be seen as sets of irregularly spaced points in a high-dimensional space. The weights of the kernel are associated with the offsets of these surrounding points with respect to the center point, and the convolution operation is performed by sliding the kernel over the input point cloud, multiplying the weights of the kernel with the corresponding features of the surrounding points, and summing the products. This process is repeated at each location of the point cloud, resulting in a new set of features that represent the convolved output.

Pointwise-CNN [108] employs a unique approach to define convolutional kernels on each grid by transforming non-uniform 3D point clouds into uniform grids, with weights assigned to all points that fall within the same grid. The output of the current layer is determined by computing the mean features of all the nearby points on the same grid, which are weighted and aggregated from all the grids. Meanwhile, Mao et al. [109] introduced the interpolated convolution operator InterpConv to assess the geometric relations between input point clouds and kernel-weight coordinates by superimposing point features onto neighboring discrete convolutional kernel-weight coordinates.

To achieve rotation invariance, Zhang et al. [110] introduced the RIConv operator, which transforms convolution into 1D using a clustering approach on low-level rotation invariant geometric features. Another approach proposed by Zhang et al. [111] is shellConv, an efficient permutation invariant convolution for point cloud deep learning. It partitions the local point neighborhood into concentric spherical shells, extracting representative features based on the statistics of the points inside. ShellNet [111] utilizes Shell-Conv as the core convolution, enabling it to handle larger receptive fields with fewer layers. However, it may not capture long-range point relations and overlooks certain patterns present in point cloud structures. To overcome this limitation, Point-PlaneNet [112] introduces a new neural network that leverages spatial local correlations by considering the distance between points and planes. The proposed PlaneConv operation learns a set of planes in $R^n$ space, allowing it to extract local geometric features from point clouds. Additionally, DeltaConv [113] introduces anisotropic filters on point clouds by mixing geometric operators from vector calculus, which allows the network to be split into scalar and vector streams that can expressively represent directional information.

**Continuous convolution** Current 3D convolution methods differ from traditional discrete convolution by defining convolutional kernels in a continuous space. Instead of fixed-size kernels sliding over a grid structure as in 2D convolution, these methods assign weights to neighboring points based on their spatial distribution relative to the center point. This allows for a more flexible and detailed feature extraction process, as 3D convolution can be seen as a weighted sum over a subset of points in continuous space.

In RS-CNN [115], the convolutional network is based on relation-shape convolution. The input to an RS-Conv kernel is a local subset of points around a given point. The mapping from low-level relations like Euclidean distance and relative location is learned using an MLP to high-level relations between points in the local subset. Using a collection of learnable kernel points, Thomas et al. [116] suggested both rigid and flexible Kernel Point Convolution (KPConv) operators for 3D point clouds. Liu et al. in their work DensePoint [117], described comvolution as a Single-layer Perceptron (SLP) with a nonlinear activator. To fully exploit the contextual information, features are learned by concatenating all of the previous layers' features. The convolution kernel is divided into spatial and feature components by ConvPoint [118]. The spatial part's positions are chosen at random from a unit sphere, and the weighting function is trained using a basic MLP. In PointConv [119], convolution is defined as a Monte Carlo estimation of a continuous 3D convolution with regard to an important sample. A weighting function and a density function are used in the procedure, which is accomplished using MLP layers and kernelized density estimation. The 3D convolution is further simplified into two operations: matrix multiplication and 2D convolution, in oreder to increase memory and computational performance. Its memory consumption can be lowered by 64 times with the same parameter settings.

Several methods have been proposed to handle large-scale point cloud scenes using feature fusion, such as Spider-CNN [120]. SpiderCNN uses a unit called SpiderConv that extends convolution operations on regular grids by combining a step function with a Taylor expansion defined on the k nearest neighbors. The Taylor expansion captures the inherent local geometric fluctuations by interpolating arbitrary values at the vertices of a cube, whereas the step function catches the coarse geometry by storing the local geometric distance. PCNN [121] is another 3D convolution network that utilizes the radial basis function for processing point clouds. Its point convolution operator is derived from extension operators that enable the transformation of point data into a continuous function space. SPHNet [122], which is based on PCNN [121], achieves rotation invariance by integrating spherical harmonic kernels during volumetric function convolution.

Designing efficient CNNs for point cloud analysis is a challenging task, requiring a delicate trade-off between accuracy and speed. Although CNNs have achieved remarkable success in image and pattern recognition, increasing the network complexity often results in decreased speed. This challenge is further amplified when dealing with point clouds, as they can contain a large number of points with varying densities.

Table 4 includes models from both discrete and continuous convolution methods. The results indicate that DeltaNet attained the highest overall accuracy (OA) on both the ModelNet40 and ScanObjectNN datasets. DensePoint, on the other hand, achieved the best OA on the ModelNet10 dataset. Moreover, PointConv exhibited the highest F1 score on the Intra dataset compared to other convolution-based methods.

### 4.1.3 Hierarchical methods

Hierarchical data structures like kd-trees and octrees are commonly employed in point cloud processing to construct networks. These networks represent the point cloud in a structured manner and learn features hierarchically, from leaf nodes to root nodes. By partitioning the point cloud into subsets of points at different levels of detail, these methods allow the model to capture local details at lower levels and global context at higher levels. As a result, these methods are effective in reducing the computational complexity of point cloud processing tasks.

In their paper, Lei et al. [123] introduced an octree guided CNN with spherical convolutional kernels applied to each layer corresponding to the octree layers. Compared to OctNet [124], which relies on octree data structures, Kd-Net [125] utilizes multiple K-d trees with different splitting directions, with non-leaf node representations computed using an MLP. Parameter sharing based on node the splitting type enables Kd-Net to efficiently learn hierarchical features while managing memory consumption.

To achieve feature learning and aggregation, 3DContextNet [126] utilizes a balanced K-d tree to learn and aggregate features, leveraging both local and global contextual cues. MLPs are employed to model the relationships between positions, allowing feature learning at each level. The non-leaf nodes compute features from their children nodes using MLP and max pooling, enabling classification until reaching the root node. SO-Net [127] establishes its structure through point-to-node k-nearest neighbor search and a Self-Organizing Map (SOM), ensuring permutation invariance. The SOM simulates the spatial distribution of point clouds by setting the positions of points, while individual point features are learned through fully connected layers. Pre-training with a point cloud auto-encoder is proposed to enhance network performance in various applications. However, processing large and complex scenes with this network may encounter limitations due to the massive amount of point cloud data involved.

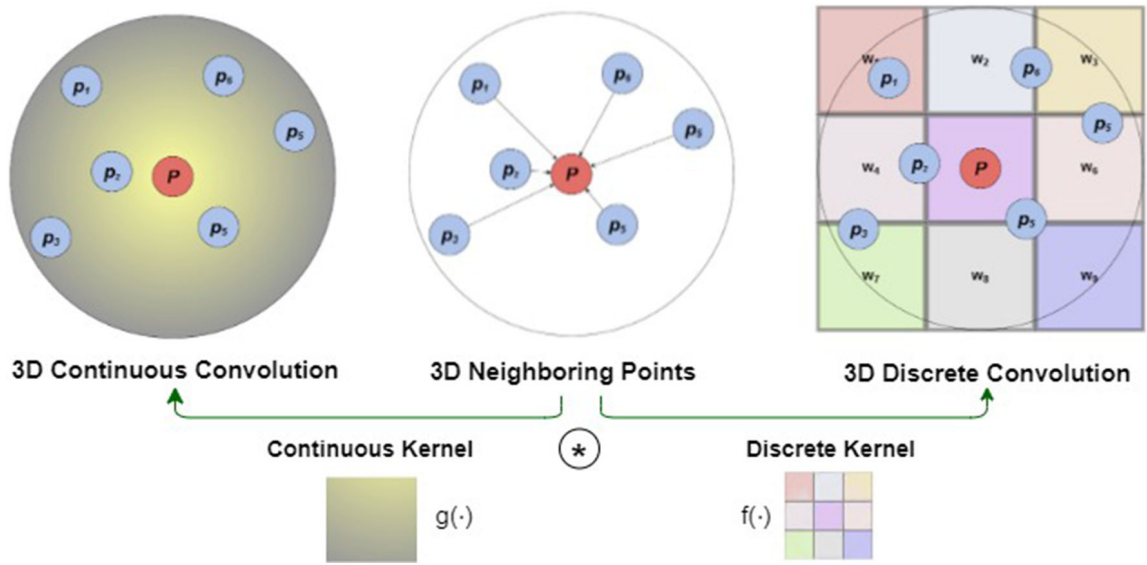

**Fig. 5** Different types of point convolution [114]

DRNet [128] is another hierarchical network that learns local point features from the point cloud in different resolutions. The DRNet architecture consists of two branches: a Full-Resolution (FR) branch and a Multi-Resolution (MR) branch. The FR branch learns local point features from the full-resolution point cloud. The MR branch learns local point features from downsampled versions of the point cloud. The two branches are then fused to produce a final feature representation.

Table 4 clearly illustrates that among the hierarchical methods, So-Net consistently outperformed all others across the assessed datasets.

### 4.1.4 Graph-based methods

Graph-based networks provide an alternative approach to analyzing point clouds by representing points as vertices in a graph connected by directed edges. These networks operate in either the spatial or spectral domain for feature learning. In the spatial domain, MLP-based convolutions are applied to spatial neighbors, and pooling generates coarsened graphs by aggregating neighboring features. In the spectral domain, convolutions are achieved through spectral filtering using the eigenvectors of the graph Laplacian matrix [129, 130]. Each vertex is assigned features like coordinates, intensities, or colors, while geometric properties between connected points are assigned to edges. Numerous graph-based approaches have been proposed for point cloud analysis, each with its unique method of generating and manipulating graphs in the feature space.

PointWeb [131], based on PointNet++ [4], uses Adaptive Feature Adjustment (AFA) to improve point features in the local neighborhood context, generating a graph in the feature space that is dynamically modified after each layer. DGCNN [132] also generates a graph in the feature space, and an MLP is used for feature learning for each edge in EdgeConv's core layer. Channel-wise symmetric aggregation is used for edge features associated with each point's neighbors. In addition, LDGCNN [133] improves the performance of DGCNN [132] by removing the transformation network and linking the hierarchical features from different layers.

Liu et al. [134] presented a Dynamic Points Agglomeration Module (DPAM) based on graph convolution to reduce the process of point agglomeration that includes sampling, grouping, and pooling to a single step. This is accomplished by multiplying the agglomeration matrix and the points feature matrix. A hierarchical learning architecture is built by stacking multiple DPAMs based on the PointNet architecture. DPAM dynamically exploits the relationship between points and agglomerates points in a semantic space, as opposed to PointNet++'s hierarchical methodology [4]. On the other hand, KCNet [135] takes a different approach by learning features based on kernel correlation to exploit local geometric structures. By defining kernels as a collection of learnable points, KCNet characterizes the geometric types of local structures, and subsequently determines the affiliation between the kernel and a specific point's neighborhood.

In RGCNN [136], a graph is built by linking each point in the point cloud to all other points and updates the graph

Laplacian matrix in each layer. The loss function includes a graph-signal smoothness prior to improve the comparability of features among nearby vertices. Alternatively, in PointGCN [137], a graph is constructed from a point cloud using k nearest neighbors, and each edge is weighted using a Gaussian kernel. The graph spectral domain is utilized to design convolutional filters using Chebyshev polynomials. To capture both global and local properties of the point cloud, global pooling and multi-resolution pooling techniques are employed. Graph convolutional networks (GCN) surpass other point-based models by preserving data granularity and utilizing point interconnectivity. However, data structure operations such as Farthest Point Sampling (FPS) and neighbor point querying consume a significant amount of time in point-based networks, limiting their speed and scalability.

To address this issue, Xu et al. [9] introduced Grid-GCN, a fast and scalable method for point cloud learning. Grid-GCN utilizes Coverage-Aware Grid Query (CAGQ), a data structuring technique that enhances spatial coverage and reduces theoretical temporal complexity by leveraging grid space efficiency. CAGQ achieves a 50% speedup compared to common sampling methods like FPS and Ball Query.

Additionally, Yang et al. [153] proposed PointManifold, a point cloud classification method based on graph neural networks and manifold learning. PointManifold employs various learning algorithms to embed point cloud features, enhancing the assessment of geometric continuity on the surface. By acquiring the point cloud nature in a low-dimensional space and combining it with features in the original 3D space, the representation capabilities and classification network performance are improved.

In [154], a novel method called Convolution in the Cloud (CIC) is proposed for learning deformable kernels in 3D graph convolution networks. CIC involves dynamically deforming a cloud of kernels to match the local structure of the point cloud. It consists of two stages: randomly sampling initial kernels and iteratively updating them based on a loss function that measures the discrepancy with the ground truth label. Meanwhile, Xu et al.'s Position Adaptive Convolution (PAConv) [155] presents a generic convolution procedure for 3D point cloud analysis. PAConv dynamically builds convolution kernels using self-adaptively learned weight matrices from point positions via the ScoreNet module. This data-driven approach allows PAConv to handle irregular and unordered point cloud data more effectively than traditional 2D convolutions. CurveNet, a proposition by Xiang et al. [156], enhances point cloud geometry learning through a novel aggregation strategy. CurveNet utilizes a curve grouping operator and a curve aggregation operator to generate continuous sequences of point segments and effectively learn features.

Table 4 presents a comparative analysis of multiple graph-based techniques for 3D point cloud classification. Notably, CurveNet [156] demonstrated remarkable performance with the highest OA of 94.20% on the ModelNet 40 dataset, outshining other graph-based methods. Meanwhile, Grid-GCN [9] demonstrated exceptional performance by securing the top OA and mAcc on the ModelNet 10 dataset among all methodologies evaluated.

### 4.1.5 Recurrent neural network-based methods

RNNs are popular for processing temporal data and have been applied in point cloud analysis to capture local context. These neural networks utilize their internal state to handle variable length sequences of inputs, making them well-suited for point cloud data. Various RNN-based techniques have been developed, highlighting the significance of local context in point cloud analysis.

RCNet [157] constructs a permutation-invariant network for 3D point cloud processing using regular RNN and 2D CNN. After partitioning the point cloud into parallel beams and sorting them along a specified dimension, each beam is input into a shared RNN. For hierarchical feature aggregation, the learnt features are used as an input to an efficient 2D CNN. RCNet-E is proposed to ensemble multiple RCNets with varied partitions and sorting directions to improve its description ability. Another RNN-based model, Point2Sequence [158], identifies correlations between distinct locations in local point cloud regions. To aggregate local region features, it treats features learnt from a local region at many scales as sequences and feeds these sequences from all local regions into an RNN-based encoder-decoder structure. Several other methods also learn from both 3D point clouds and 2D images.

According to Table 4, Point2sequence achieves the highest overall accuracy on the ModelNet 40 dataset, while RCNet-E performs best on the ModelNet 10 dataset over all other RNN-based methods.

### 4.1.6 Transformer-based methods

One of the most significant recent breakthroughs in natural language processing and 2D vision is the Transformer [209], which has demonstrated superior performance in capturing long-range relationships. The success of Transformer has also led to notable improvements in point-based models through the use of self-attention. With the attention mechanism, the Transformer can weigh the relevance of each point to the others, enabling better feature extraction and discrimination. The development of Transformer-based architectures [160, 162] has greatly enhanced performance. Nevertheless, the bottleneck of these models still remains

**Table 4** Comperative 3D point cloud classification result on various available datasets for Point based methods

| Model name | Year | Input | Params (in Million) | ModelNet 40 (OA) | (mAcc) | ModelNet 10 (OA) | (mAcc) | ScanObjectNN (OA) | (mAcc) | Intra (F1) |
|---|---|---|---|---|---|---|---|---|---|---|
| **Pointwise MLP methods** | | | | | | | | | | |
| DeepSets [138] | 2017 | PC | – | 82.00% | – | – | – | – | – | – |
| PointNet [3] | 2017 | PC | 3.5 | 89.20% | 86.20% | – | – | 68.20% | 63.40% | 68.40% |
| PointNet++ [4] | 2017 | PC+PF | 1.74 | 90.70% | – | – | – | 77.90% | 75.40% | **90.30%** |
| MO-Net [100] | 2019 | PC | 3.1 | 92.40% | 90.30% | – | – | – | – | – |
| SRN-PointNet++ [97] | 2019 | PC | – | 91.50% | – | – | – | – | – | – |
| JUSTLOOKUP [102] | 2019 | PC | – | 89.50% | 86.40% | 92.90% | 92.10% | – | – | – |
| HPGCNN+DC [139] | 2020 | PC | 1.0 | 92.60% | 90.40% | – | – | – | – | – |
| PointASNL [7] | 2020 | PC+PF | 3.98 | 93.20% | – | **95.90%** | – | – | – | – |
| ASSANet [140] | 2021 | PC | – | 92.90% | – | – | – | – | – | – |
| RPNet-W9 [103] | 2021 | PC+PF | – | 94.10% | – | – | – | – | – | – |
| PointMLP [104] | 2022 | PC | 12.6 | 94.50% | 91.40% | – | – | 85.40% | 83.90% | – |
| PointNeXt-S [105] | 2022 | PC | 1.4 | 93.20% | 90.80% | – | – | 87.70% | 85.80% | – |
| PointNeXt+HyCoRe [141] | 2022 | PC | – | – | 87.60% | – | – | 88.30% | 87.00% | – |
| PointMLP+HyCoRe [141] | 2022 | PC | – | 94.50% | 91.90% | – | – | 87.20% | 85.90% | – |
| PointStack [106] | 2022 | PC | 1.62 | **94.70%** | **92.40%** | – | – | **89.40%** | **88.50%** | – |
| **Hierarchical methods** | | | | | | | | | | |
| KD-Net [125] | 2017 | PC | 2.0 | 91.80% | 88.50% | 94.00% | 93.50% | – | – | – |
| SO-Net [127] | 2018 | PC+Normals | – | **93.40%** | **90.80%** | **96.70%** | **95.50%** | – | – | **86.80%** |
| SCN [142] | 2018 | PC | – | 90.00% | 87.60% | – | – | – | – | – |
| A-SCN [142] | 2018 | PC | – | 89.80% | 87.40% | – | – | – | – | – |
| 3DContextNet [126] | 2018 | PC | – | 90.20% | – | – | – | – | – | – |
| 3DContextNet [126] | 2018 | PC+Normals | – | 91.10% | – | – | – | – | – | – |
| **Convolution-based methods** | | | | | | | | | | |
| SphericalCNNs [143] | 2018 | PC | 0.5 | 86.90% | – | – | – | – | – | – |
| Pointwise-CNN [108] | 2018 | PC+PF | – | 86.10% | 81.40% | – | – | – | – | – |
| MCConvolution [144] | 2018 | PC | – | 90.90% | – | – | – | – | – | – |
| SpiderCNN [120] | 2018 | PC+PF | – | 92.40% | – | – | – | 73.70% | 69.80% | 87.20% |
| PointCNN [145] | 2018 | PC | 0.45 | 92.20% | 88.10% | – | – | 78.50% | 75.10% | 87.50% |

**Table 4** continued

| Model name | Year | Input | Params (in Million) | ModelNet 40 (OA) | (mAcc) | ModelNet 10 (OA) | (mAcc) | ScanObjectNN (OA) | (mAcc) | Intra (F1) |
|---|---|---|---|---|---|---|---|---|---|---|
| Flex-Convolution [146] | 2018 | PC | – | 90.20% | – | – | – | – | – | – |
| PCNN [121] | 2018 | PC | 1.4 | 92.30% | – | 94.90% | – | – | – | – |
| PointConv [119] | 2019 | PC+PF | 18.6 | 92.50% | – | – | – | – | – | **88.30%** |
| RS-CNN [115] | 2019 | PC | – | 93.60% | – | – | – | – | – | – |
| GeoCNN [147] | 2019 | PC | – | 93.40% | 91.10% | – | – | – | – | – |
| Ψ-CNN [123] | 2019 | PC | – | 92.00% | 88.70% | 94.60% | 94.40% | – | – | – |
| A-CNN [148] | 2019 | PC | – | 92.60% | 90.30% | 95.50% | **95.30%** | – | – | – |
| SFCNN [149] | 2019 | PC+PF | – | 92.30% | – | – | – | – | – | – |
| DensePoint [117] | 2019 | PC+PF | 0.53 | 93.20% | – | **96.60%** | – | – | – | – |
| KPConvrigid [116] | 2019 | PC | 15.2 | 92.90% | – | – | – | – | – | – |
| KPConvdeform [116] | 2019 | PC | – | 92.70% | – | – | – | – | – | – |
| InterpCNN [109] | 2019 | PC+PF | 12.8 | 93.00% | – | – | – | – | – | – |
| ShellNet(SS=16) [111] | 2019 | PC | – | 93.10% | – | – | – | – | – | – |
| ConvPoint [118] | 2020 | PC | – | 91.80% | 88.50% | – | – | – | – | – |
| Point-PlaneNet [112] | 2020 | PC+PF | – | 92.10% | 90.50% | – | – | – | – | – |
| DRNet [128] | 2021 | PC+PF | – | 93.10% | – | – | – | 80.30% | **78.00%** | – |
| DeltaNet [113] | 2022 | PC | – | **93.80%** | **91.20%** | – | – | **84.70%** | – | – |
| *Graph-based methods* | | | | | | | | | | |
| ECC [150] | 2017 | PC | – | 87.40% | 83.20% | 90.80% | 90.00% | – | – | – |
| KCNet [135] | 2018 | PC | 0.9 | 91.00% | – | 94.40% | – | – | – | – |
| LocalSpecGCN [151] | 2018 | PC+Normals | – | 92.10% | – | – | – | – | – | – |
| RGCNN [136] | 2018 | PC+Normals | 2.24 | 90.50% | 87.30% | – | – | – | – | – |
| 3DTI-Net [152] | 2018 | PC | 2.6 | 91.70% | – | – | – | – | – | – |
| PointGCN [137] | 2018 | PC | – | 89.51% | 86.05% | 91.91% | 91.57% | – | – | – |
| PointWeb [131] | 2019 | PC | – | 92.30% | 89.40% | – | – | – | – | – |
| DGCNN [132] | 2019 | PC | 1.84 | 93.50% | 90.70% | – | – | 78.10% | 73.60% | 73.80% |
| LDGCNN [133] | 2019 | PC | 1.08 | 92.90% | 90.30% | – | – | – | – | – |
| DPAM [134] | 2020 | PC+Normals | – | 91.90% | 89.90% | 94.60% | 94.30% | – | – | – |
| Grid-GCN [9] | 2020 | PC | – | 93.10% | **91.30%** | **97.50%** | **97.40%** | – | – | – |
| PointManifold [153] | 2020 | PC | – | 93.00% | 90.10% | – | – | – | – | – |
| 3D-GCN [154] | 2020 | PC | 0.89 | 92.10% | – | – | – | – | – | – |
| PAConv [155] | 2021 | PC | – | 93.90% | – | – | – | – | – | **90.60%** |
| CurveNet [156] | 2021 | PC | – | **94.20%** | – | 96.30% | – | – | – | – |

**Table 4** continued

| Model name | Year | Input | Params (in Million) | ModelNet 40 (OA) | (mAcc) | ModelNet 10 (OA) | (mAcc) | ScanObjectNN (OA) | (mAcc) | Intra (F1) |
|---|---|---|---|---|---|---|---|---|---|---|
| *RNN-based Methods* | | | | | | | | | | |
| RCNet [157] | 2019 | PC | 13.3 | 91.60% | – | 94.70% | – | – | – | – |
| RCNet-E [157] | 2019 | PC | 39.9 | 92.30% | – | **95.60%** | – | – | – | – |
| Point2sequence [158] | 2019 | PC | – | 92.60% | **90.40%** | 95.30% | **95.10%** | – | – | – |
| *Transformer-based methods* | | | | | | | | | | |
| PAT [159] | 2019 | PC | – | 91.70% | – | – | – | – | – | – |
| Point Transformer [160] | 2021 | PC | – | 93.70% | 90.60% | – | – | – | – | – |
| PCT [161] | 2021 | PC+Normals | 2.88 | 93.20% | – | – | – | – | – | 91.40% |
| Point Transformer [162] | 2021 | PC+Normals | 13.5 | 92.80% | – | – | – | – | – | – |
| 3DMedPT [98] | 2021 | PC+Normals | 1.54 | 93.40% | – | – | – | – | – | **93.60%** |
| Perceiver [163] | 2021 | PC | – | 85.70% | – | – | – | – | – | – |
| PointTnT [164] | 2022 | PC | 3.9 | 92.60% | – | – | – | 85.00% | 83.50% | – |
| Patchformer [165] | 2022 | PC | 2.45 | 93.50% | – | – | – | – | – | – |
| PTv2 [166] | 2022 | PC | – | **94.20%** | 91.60% | – | – | – | – | – |
| LCPFormer [167] | 2023 | PC+Normals | – | 93.60% | 90.70% | – | – | – | – | – |
| SPoTr [168] | 2023 | PC | – | – | 91.00% | – | – | **88.60%** | **86.80%** | – |
| IBT [169] | 2023 | PC | – | 93.60% | – | – | – | 82.80% | 80.00% | – |
| APES(global-based) [170] | 2023 | PC | – | 93.80% | – | – | – | – | – | – |
| *Other methods* | | | | | | | | | | |
| DeepRBFNet [171] | 2018 | PC | 3.2 | 90.2% | 87.80% | – | – | – | – | – |
| DeepRBFNet [171] | 2018 | PC+Normals | 3.2 | 92.1% | 88.80% | – | – | – | – | – |
| PointHop [172] | 2019 | PC | – | 89.10% | 84.40% | – | – | – | – | – |
| FG-Net [173] | 2020 | PC | – | 93.80% | **93.10%** | – | – | – | – | – |
| PointHop++ [174] | 2020 | PC | 0.16 | 91.10% | 87.00% | – | – | – | – | – |
| PRA-Net [175] | 2021 | PC | – | 93.70% | 91.20% | – | – | 82.10% | 79.10% | – |
| GDANet [176] | 2021 | PC | – | 93.80% | – | – | – | **88.50%** | – | – |
| PointSCNet [177] | 2021 | PC + Normals | 1.827 | 93.70% | – | – | – | – | – | – |
| APP-Net [178] | 2022 | PC + Normals | 1.827 | **94.00%** | – | – | – | 87.00% | – | – |
| PointMetaBase-XXL [179] | 2023 | PC | 22.7 | – | – | – | – | 87.90% | – | – |
| *Unsupervised-based methods* | | | | | | | | | | |
| FoldingNet [180] | 2018 | PC | – | 88.40% | – | 94.40% | – | – | – | – |
| PPF-FoldNet [181] | 2018 | PC+Normals | – | – | – | – | – | – | – | – |

Table 4 continued

| Model name | Year | Input | Params (in Million) | ModelNet 40 (OA) | (mAcc) | ModelNet 10 (OA) | (mAcc) | ScanObjectNN (OA) | (mAcc) | Intra (F1) |
|---|---|---|---|---|---|---|---|---|---|---|
| Latent-GAN [182] | 2018 | PC | – | 84.50% | – | **95.40%** | – | – | – | – |
| MRTNet-VAE [183] | 2018 | PC | – | 86.40% | – | – | – | – | – | – |
| Hassani et al. [184] | 2019 | PC | – | 89.10% | – | – | – | – | – | – |
| 3DPointCapsNet [185] | 2019 | PC | – | 89.30% | – | – | – | – | – | – |
| Cluster-Net [186] | 2019 | PC | – | 88.80% | – | 93.80% | – | – | – | – |
| GBNet [187] | 2019 | PC | 8.39 | 93.80% | 91.00% | – | – | 80.50% | 77.80% | – |
| PointGrow [188] | 2020 | PC | 0.25 | 85.80% | – | – | – | – | – | – |
| ParAE [189] | 2021 | PC | – | 91.60% | – | – | – | – | – | – |
| OcCo+DGCNN [190] | 2021 | PC | – | 93.00% | – | – | – | 83.90% | – | – |
| OGNet + MD [191] | 2021 | PC | – | 93.31% | 90.71% | – | – | – | – | – |
| DGCNN+MD [191] | 2021 | PC | – | 93.39% | 90.26% | – | – | – | – | – |
| MD [191] | 2022 | PC | – | 93.30% | 90.26% | – | – | – | – | – |
| STRL + DGCNN [192] | 2021 | PC | – | 90.90% | – | – | – | – | – | – |
| IAE + DGCNN [193] | 2022 | PC | – | 94.20% | **91.60%** | – | – | – | – | – |
| AL-GAN + PointNet++ [194] | 2022 | PC | – | 92.70% | – | – | – | – | – | – |
| Point-MAE [195] | 2022 | PC | – | 93.80% | – | – | – | 85.20% | – | – |
| Point-BERT [196] | 2022 | PC | – | 93.80% | – | – | – | 83.07% | – | – |
| P2P [197] | 2022 | PC | 1.2 | 94.00% | **91.60%** | – | – | 89.30% | **88.50%** | – |
| PointCaps [198] | 2022 | PC+Normals | 3.52 | 94.70% | – | 91.70% | – | – | – | – |
| MAE3D [199] | 2022 | PC | – | 93.40% | – | – | – | 86.20% | – | – |
| Point-M2AE [200] | 2022 | PC | – | 94.00% | – | – | – | 86.43% | – | – |
| Hao et al. [201] | 2022 | PC | 0.88 | 94.20% | – | – | – | 82.60% | – | – |
| MaskPoint [202] | 2022 | PC | – | 93.80% | – | – | – | 84.60% | – | – |
| ACT [203] | 2022 | PC | – | 92.69% | – | – | – | 88.21% | – | – |
| I2P-MAE [204] | 2022 | PC | – | 93.72% | – | – | – | 90.11% | – | – |
| PointGPT-L [205] | 2023 | PC(1k P) | – | 94.70% | – | – | – | **93.40%** | – | – |
| PointGPT-L [205] | 2023 | PC(8k P) | – | **94.90%** | – | – | – | – | – | – |
| point2vec [206] | 2023 | PC | – | 94.80% | – | – | – | 87.50% | – | – |
| ReCoN [207] | 2023 | PC(1K P) | – | 93.90% | – | – | – | 90.65% | – | – |
| ReCoN [207] | 2023 | PC(8K P) | – | 94.20% | – | – | – | – | – | – |
| PointMLP+ULIP [208] | 2023 | PC | – | 93.30% | 89.60% | – | – | 86.90% | 85.80% | – |

Here 'Params' represents the number of parameters, while 'OA' refers to the overall accuracy, 'mAcc' denotes mean accuracy, and 'F1' stands for the Dice score. The symbol '–' means the results are unavailable. The methods are arranged in chronological order within their corresponding categories. The top-performing methods in each category have been highlighted in bold, while the method(s) achieving the best overall performance across all categories are underlined

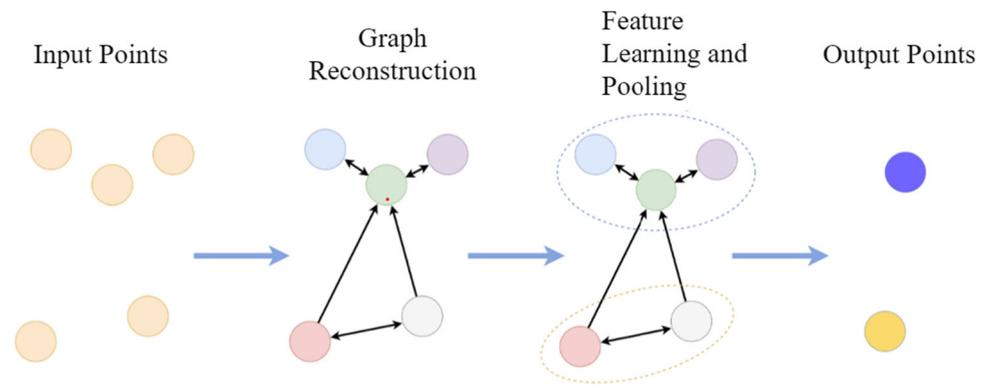

**Fig. 6** Illustration of a graph-based network

the time-consuming operation of sampling and aggregating characteristics from irregular sites.

Point Attention Transformers (PATs) [159] learns high-dimensional features by encoding each point's absolute and relative positions with respect to its neighbors. To extract hierarchical features, it utilizes a trainable, permutation-invariant, and non-linear end-to-end Gumbel Subset Sampling (GSS) layer, which captures relationships between points using Group Shuffle Attention (GSA). This unique approach enables the model to capture local structures within each group while also considering the global context of the entire point cloud. Zhao et al. [160] proposed a similar model, the Point Transformer, which employs a self-attention module to retrieve spatial characteristics from local neighborhoods around each point, and encodes positional information. The network has a highly expressive Point Transformer layer, which is invariant to permutation and cardinality, making it ideal for point cloud processing.

Point Transformer V2 [166] is an enhanced version of the Point Transformer architecture for 3D point cloud processing. It introduces two innovations: grouped vector attention and partition-based pooling. Grouped vector attention reduces computational cost by performing attention only within groups of points, maintaining accuracy while learning long-range dependencies. Partition-based pooling improves accuracy on large point clouds by dividing them into smaller partitions and pooling features within each partition, enabling global feature learning with reduced computational load. Engel et al. [162] introduced another model called Point Transformer, which operates directly on unordered and unstructured point sets. The Point Transformer uses a local–global attention mechanism to capture spatial point relations and shape information, allowing it to extract both local and global aspects of the point cloud. SortNet, a component of Point Transformer, produces input permutation invariance by selecting points based on a learned score. The Point Transformer produces a sorted and permutation invariant feature list that can be utilized directly in standard computer vision applications.

Perceiver, another attention-based architecture introduced in [163] is a scalable attention-based architecture for high-dimensional inputs, such as images, movies, and audio, without domain-specific assumptions. It utilizes cross-attention and latent self-attention blocks to process a fixed-dimensional latent bottleneck. 3D medical point Transformer (3DMedPT) [98] is an attention-based model specifically designed for medical point clouds for examining the complex biological structures that are vital for disease detection and treatment. Insufficient training samples of medical data can lead to poor feature learning. To enhance feature representations in medical point clouds, it employs an attention module to capture local and global feature interactions, position embeddings for precise local geometry, and Multi-Graph Reasoning (MGR) for global knowledge transmission.

Similarly, Berg et al. [164] propose the two-stage Point Transformer-in-Transformer (Point-TnT) technique, which combines both local and global attention mechanisms by producing patches of local features via a sparse collection of anchor points. Self-attention can then be used on both the points within the patches and the patches themselves, resulting in a highly effective method for processing unstructured point cloud data. LCPFormer [167] is a recent transformer-based architecture for 3D point cloud analysis. LCPFormer introduces a novel local context propagation (LCP) module that enables the model to learn long-range dependencies between points in a point cloud. The LCP module works by first dividing the input point cloud into local regions. Then, it propagates the features of each local region to its neighboring local regions. This allows the model to learn long-range dependencies between points that are not directly connected.

To learn local and global shape contexts with reduced complexity Park et al. introduced SPoTr [168], a self-positioning mechanism that works by first randomly selecting a subset of points from the input point cloud. These points are then used to create a local coordinate system. The remaining points are then projected into this local coordinate system. This allows the model to learn local shape contexts without the need for global attention. Wu et al. [170] proposed an Attention-Based Point Cloud Edge Sampling (APES) for sampling points from a point cloud based on their importance to the outline of the object. The attention mechanism in APES is based on the self-attention mechanism used in transformer models. The self-attention mechanism computes the attention weights between each point in the point cloud and all other points in the point cloud. The points with the highest weights are then selected to form a new, downsampled point cloud.

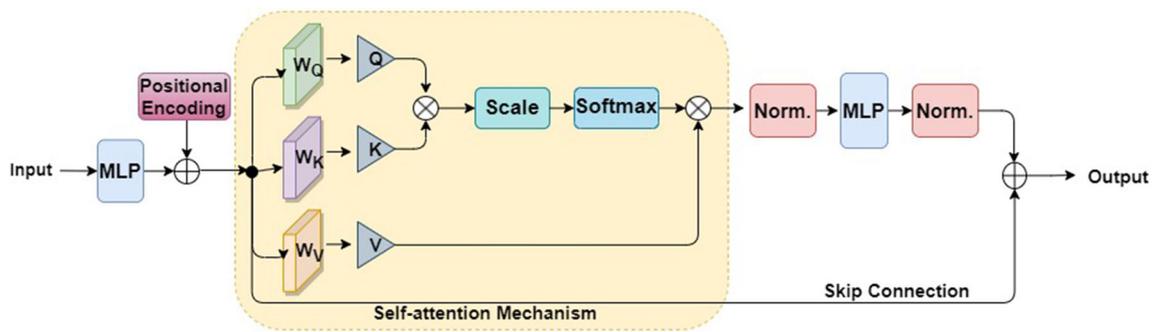

**Fig. 7** Illustration of the transformer based encoder architecture [210]

Table 4 presents a comparison of various pointwise transformer-based methods on different datasets. Here, PTv2 [166] achieved the highest OA and mAcc on the ModelNet 40 dataset, while SPoTr [168] showed best performance on ScanObjectNN dataset.

### 4.1.7 Other methods

Apart from the methods discussed earlier, there are several techniques that cannot be neatly categorized into a specific class. These methods utilize multiple modalities to learn intricate representations of point clouds, thereby enabling them to capture intricate patterns and relationships. Hence, in this section, we will explore these unconventional methods that transcend traditional classification boundaries, providing a comprehensive overview of each.

With prior knowledge of kernel positions and sizes, RBFNet [171] aggregates features from sparsely distributed Radial Basis Function (RBF) kernels to explicitly characterize the spatial distribution of points. PointAugment, an auto-augmentation framework introduced by Li et al. [211], optimizes and augments point cloud data by automatically learning each input sample's shape-wise transformation and pointwise displacement. Prokudin et al. [212] transform the point cloud into a vector with a short fixed length by encoding the point cloud as minimal distances to a uniformly distributed basis point set sampled from a unit ball. Finally, common machine learning techniques are applied to produce the encoder representation. Cheng et al. [175] present the Point Relation-Aware Network (PRA-Net), comprising two modules: intra-region structure learning (ISL) and inter-region relationship learning (IRL). The ISL module can adaptively incorporate local structural information into point features, while the IRL module dynamically and effectively preserves inter-region relations using a differentiable region partition method and a representative point-based strategy.

FG-Net [173] proposes a comprehensive deep learning framework for large-scale point cloud understanding that achieves accurate and real-time performance with a single GPU. The network incorporates a noise and outlier filtering mechanism, utilizes a deep CNN to exploit local feature correlations and geometric patterns, and employs efficient techniques such as inverse density sampling and feature pyramid-based residual learning to address efficiency concerns. Another recent work in this area is proposed by Xu et al. in GDANet [176]. It introduces the Geometry-Disentangled Attention Network, which dynamically disentangles point clouds into contour and flat parts of 3D objects. It utilizes the disentangled components to generate holistic representations and applies different attention mechanisms to fuse them with the original features. The network also captures and refines 3D geometric semantics from the disentangled components to supplement local information.

PointSCNet [177] captures the geometrical structure and local region correlation of a point cloud using three key components: a space-filling curve-guided sampling module, an information fusion module, and a channel-spatial attention module. The sampling module selects points with geometrical correlation using Z-order curve coding. The information fusion module combines structure and correlation information through a correlation tensor and skip connections. The channel-spatial attention module enhances critical sites and feature channels for improved network representation. Lu et al. [178] proposed APP-Net, a network that utilizes auxiliary points and push and pull operations to efficiently classify point cloud data. The auxiliary points guide the network's attention to important regions, while the push and pull operations allow for efficient computation and improved feature representation.

PointMeta [179] by Lin et al. is a unified meta-architecture for point cloud analysis. It abstracts the computation pipeline into four meta-functions: neighbor update, neighbor aggregation, point update, and position embedding. These functions enable learning of local and global features, point refinement, and encoding of spatial relationships. PointMeta offers flexibility and efficiency in designing point cloud analysis models. However, a detailed computational complexity analysis is not provided in the paper.

Table 4 shows that among the models in other methods, APP-Net achieved the highest overall accuracy (OA) score of 94.00% on the ModelNet 40 dataset. However, among all the models across different methodologies, FG-Net emerged as the leader in mean accuracy (mAcc) with a score of 93.10%. On the ScanObjectNN dataset, PRA-Net achieved the highest mAcc score, and GDANet achieved the highest OA score.

## 4.2 Unsupervised training

Unsupervised representation learning is a technique that aims to learn useful and informative features from unlabeled data. In the context of point cloud understanding, this approach involves training deep neural networks to extract latent features from raw, unannotated point cloud data. Unsupervised representation learning for point clouds has gained significant attention in recent years due to its ability to reduce the need for labeled data and improve the performance of various point cloud applications including natural language understanding [213], object detection [214], graph learning [215], and visual localization [216]. By pre-training deep neural networks on unlabeled data, unsupervised learning uncovers latent features without human-defined annotations, reducing reliance on labeled data. It can be categorized into generative modeling, where synthetic point clouds are generated, and self-supervised learning, which involves predicting missing information from partially observed point clouds. This active research field holds promise for improving the accuracy and efficiency of point cloud processing tasks.

### 4.2.1 Generative model-based methods

Unsupervised approaches like generative adversarial networks (GANs) [217] and autoencoders (AEs) [184] learn representation of provided data [121]. AEs consist of an encoder, internal representation, and decoder, and are widely used for data representation and generation. They can capture point cloud irregularities and address sparsity during upsampling. GANs, on the other hand, consist of a generator and discriminator, aiming to generate realistic data samples. GANs learn to produce new data with similar statistics as the training set.

FoldingNet [180] is an end-to-end unsupervised deep autoencoder network that uses the concatenation of a vectorized local covariance matrix and point coordinates as its input. Hassani and Haley [184] suggested an unsupervised multi-task autoencoder to learn point and shape features, inspired by Inception module [218] and DGCNN [132]. Multi-scale graphs are used to build the encoder. The decoder is built utilizing three unsupervised tasks: clustering, self-supervised classification, and reconstruction, all of which are combined and trained together with a multi-task loss.

Latent-GAN [182] is one of the first networks to use GAN for raw point clouds. The authors discuss various methods such as autoencoders, variational autoencoders (VAE) [219], GAN, and flow-based models that have been proposed for learning effective representations of 3D point clouds and generating new ones. 3DAAE [220] can learn the representation of 3D point clouds by using an end-to-end approach. This model generates output by first learning a latent space for 3D shapes and then using adversarial training. The inventors of 3DAAE created a 3D autoencoder that takes 3D data as input and produces a 3D output.

3D point-capsule networks [122] have been developed to address the sparsity issue in point clouds while preserving their spatial arrangements. This network extends the 2D capsule networks to the 3D domain and uses an autoencoder to handle the sparsity of point clouds. In contrast, 3DPointCapsNet [185] incorporates pointwise MLP and convolutional layers to extract point-independent features and employs several maxpooling layers to derive a global latent representation. Unsupervised dynamic routing is then used to learn representative latent capsules. In addition, Pang et al. [195] introduced a novel approach using masked autoencoders for self-supervised learning of point clouds. They addressed challenges related to point cloud properties, such as location information leakage and uneven density, by dividing the input into irregular patches, applying random masking, and using an asymmetric design with shifting mask token operation. This enabled a transformer-based autoencoder to learn latent characteristics from unmasked patches and reconstruct the masked ones.

Point clouds are discrete samples of a continuous three-dimensional surface. As a result, sample differences in the underlying 3D shapes are inescapable. The conventional autoencoding paradigm requires the encoder to record sampling fluctuations in the same way that the decoder must recreate the original point cloud. Yan et al. [193] introduced the Implicit Autoencoder (IAE) to overcome the challenge of sampling fluctuations in point clouds. By using an implicit decoder instead of a point cloud decoder, IAE generates a continuous representation that can be shared across multiple samplings of the same model. This approach allows the encoder to focus on learning valuable features by ignoring sampling changes during reconstruction.

Point-BERT [196] is a more advanced version of BERT that employs transformers to generalize 3D point cloud

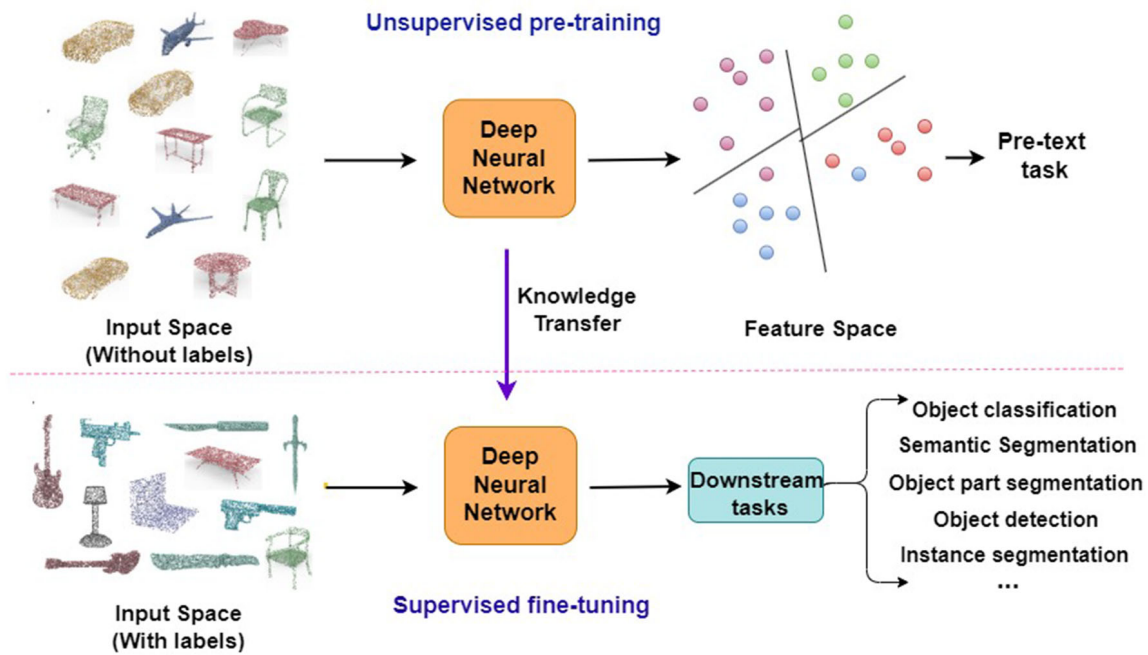

**Fig. 8** The general pipeline of unsupervised representation learning on point clouds. Neural networks are trained on unannotated point clouds using unsupervised learning, followed by transfer of learned representations to downstream tasks for network initialization. Pre-trained networks can then be fine-tuned with a small amount of annotated task-specific point cloud data [221]

learning. A point cloud tokenizer with a discrete Variational AutoEncoder (dVAE) is intended to generate discrete point tokens containing significant local information once the network separates a point cloud into many local point patches. Then it feeds some patches of input point clouds into the backbone transformers, using random masking. Under the supervision of point tokens obtained by the tokenizer, the pre-training goal is to recover the original point tokens at the masked places. In [200], Zhang et al. introduced Point-M2AE, a pre-training framework for learning 3D representations of point clouds. It utilizes a multi-scale masking strategy, pyramid architectures, local spatial self-attention, and complementary skip connections to capture detailed information and high-level semantics of shapes. This paper also highlights the significance of a lightweight decoder in Point-M2AE, which contributes to the reconstruction of point tokens and promotes the quality of shape representation. [199] discusses another method for learning representations for 3D point clouds using masked autoencoders. In the proposed method, a portion of the points in the point cloud are masked out and the masked autoencoder is trained to reconstruct the masked out points.

To addresses the challenge of limited 3D datasets for learning high-quality 3D features, in [204], the authors proposed Image-to-Point Masked Autoencoders (I2P-MAE). By leveraging 2D pre-trained models, I2P-MAE reconstructs masked point tokens using an encoder-decoder architecture.

It employs a 2D-guided masking strategy to focus on semantically important point tokens and capture key spatial cues for significant 3D structures. Through self-supervised pre-training and multi-view 2D feature reconstruction, I2P-MAE enables superior 3D representations from 2D pre-trained models. In their paper [203], Dong et al. proposed ACT (Autoencoders as Cross-Modal Teachers), a method for training 3D point cloud models using pretrained 2D image transformers. ACT involves two steps: pretraining a 2D image transformer on a large image dataset and fine-tuning it on a 3D point cloud dataset. The fine-tuning process utilizes the 2D image transformer to generate a latent representation of the 3D point cloud, which is then used to train the 3D point cloud model.

### 4.2.2 Self-supervised methods

Self-supervised learning in point cloud processing is a powerful technique that leverages unannotated data to improve performance across various applications. By incorporating geometric and topological priors, models can learn feature representations. This involves training a model to predict local geometric properties, such as normals or curvatures, using the point positions as input.

Due to the complex nature of 3D scene understanding tasks and the vast differences provided by camera perspectives, illumination, occlusions, and other factors, there are yet no effective and generalizable pre-trained models available. In their paper, Huang et al. [192] address this problem by proposing a self-supervised Spatio-temporal Representation Learning (STRL) framework that learns from unlabeled 3D point clouds. STRL utilizes two temporally correlated frames, applies spatial data augmentation, and self-supervisedly learns invariant representations.

Occlusion Completion (OcCo) is an unsupervised pre-training method proposed by Wang et al. [190], which comprises of three separate mechanisms. The first step is to use view-point occlusions to create masked point clouds. The second step is to complete reconstructing the occluded point cloud, and the final step is to use the encoder weights as the initialization for the downstream point cloud task. Sun et al. [191] developed a novel self-supervised learning technique called Mixing and Disentangling (MD) for learning 3D point cloud representations in response to the enormous success of self-supervised learning. The authors combined two input shapes and demand that the model learn to distinguish the inputs from the mixed shape. This reconstruction task serves as the pretext optimization objective for self-supervised learning, and the disentangling process drives the model to mine the geometric prior knowledge.

Xue et al. [208] introduced ULIP (Unified Language-Image-Point Cloud) as a pre-training method for learning a unified representation of language, images, and point clouds in 3D understanding. ULIP learns a common embedding space for these modalities, enabling various 3D tasks. By leveraging the shared information about 3D objects, ULIP creates informative and discriminative representations. It utilizes a large-scale dataset of language, images, and point clouds, generated with triplets describing the same object, and trains the model to predict the missing modality in each triplet. PointCaps [198] introduces a capsule network, a structured representation learning approach for point clouds. The method consists of two operations: learning local and global features of the point cloud using the capsule network, and subsequently classifying the point cloud into predefined classes. Qi et al. [207] propose ReCon (Contrast with Reconstruct), a self-supervised method for 3D representation learning. ReCon combines contrastive learning and generative pretraining in two stages. In the contrastive learning step, ReCon learns local and global features of 3D point clouds through pairwise comparisons. In the generative pretraining step, ReCon learns high-level features by generating data similar to the training set

Point2vec [206] extends the data2vec [222] framework for self-supervised representation learning on point clouds, overcoming the limitation of leaking positional information during training. Point2vec unleashes the full potential of data2vec-like pre-training on point clouds. In response to the growing popularity of Large Language Models, Chen et al. introduced PointGPT [205] that extends the GPT concept to point clouds, addressing challenges such as disorder properties and low information density. PointGPT pre-trains transformer models using a point cloud auto-regressive generation task. The method employs a dual masking strategy in the extractor-generator based transformer decoder, capturing dependencies between points and generating coherent and realistic point clouds.

Table 4 presents findings encompassing both generative-based and self-supervised-based methods. The outcomes illuminate that amid all models derived from diverse methodologies, PointGPT-L secured the top OA for both the ModelNet40 and ScanObjectNN datasets.

## 5 3D point cloud semantic segmentation

The task of 3D point cloud segmentation requires a comprehensive understanding of both the overall geometric structure and the specific properties of each individual point. Depending on the level of detail required, 3D point cloud segmentation techniques can be broadly classified into three categories: semantic segmentation at the scene level, instance segmentation at the object level, and part segmentation at the part level. In this paper, our exclusive focus has been on semantic segmentation, rather than encompassing all forms of segmentation.

While many classification models have been shown to perform well on established benchmarks, they also rely on segmentation datasets to showcase their unique contributions and generalization capabilities. This section will primarily focus on models that have not been previously discussed in the classification part of this paper.

Semantic segmentation involves the partitioning of a point cloud into distinct subsets, determined by the semantic interpretation of individual points. Based on the input data representation, this segmentation can be categorized into four types, akin to the classification of 3D shapes: projection-based, discretization-based, hybrid methods, and raw point-based. Approaches such as projection [223, 224], volumetric [225, 226], and hybrid representations [227, 228] initiate the process by transforming a point cloud into an intermediary regular representation.

### 5.1 Projection-based methods

The projection-based method is a widely adopted approach for semantic segmentation of point clouds. It involves assigning semantic labels to individual points in a 3D point cloud by projecting it onto multiple 2D planes or views. Each projected view is processed using 2D segmentation techniques, and the results are fused to obtain the final semantic segmentation. This method offers advantages such as reduced complexity and the utilization of existing image-based seg-mentation techniques. However, the choice of projection views can impact segmentation accuracy, and complex point cloud geometries or non-planar surfaces may pose challenges. Nonetheless, the projection-based method remains a powerful tool for point cloud semantic segmentation. It can be further categorized into multi-View, range-view, and bird's eye view approaches.

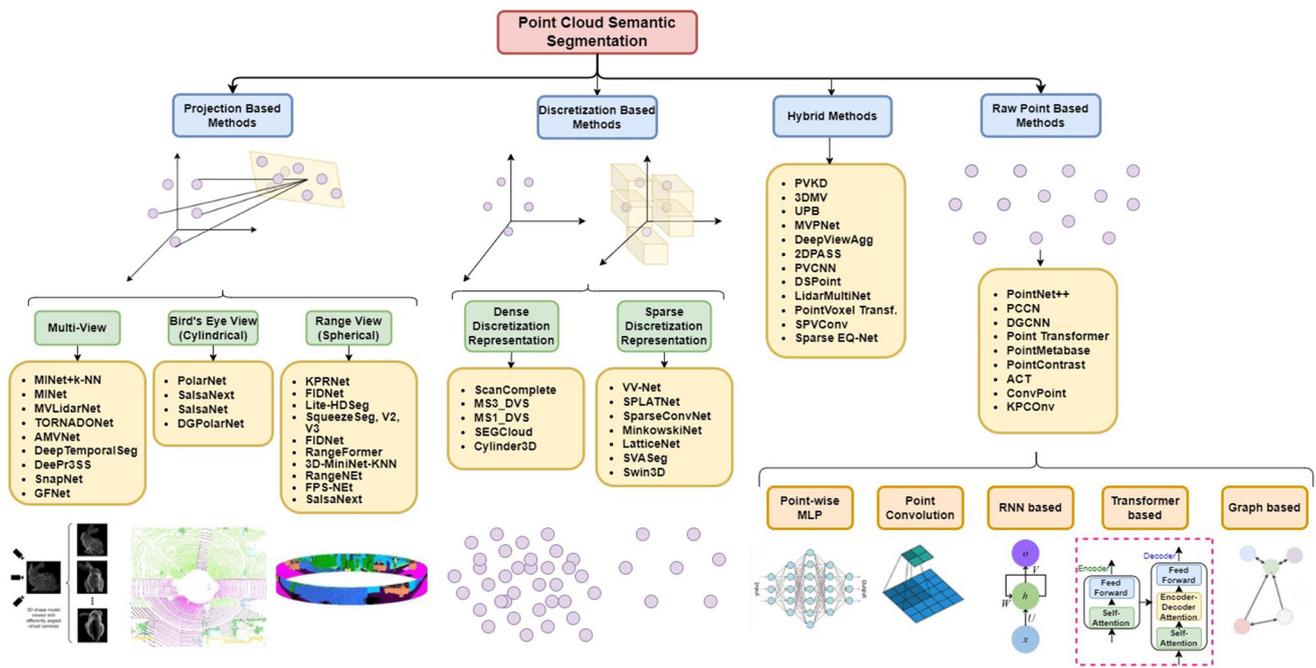

**Fig. 9** A taxonomy of deep learning methods for 3D point cloud semantic segmentation

### 5.1.1 Multi-view based method

Multi-view approaches in point cloud segmentation processing harness a potent paradigm by integrating data from diverse perspectives. This holistic strategy offers comprehensive scene insights. These methods utilizes multiple sensor viewpoints to capture a wide range of geometric details, bolstering resilience against occlusions and lighting variations. The fusion of data from multiple sources mitigates limitations tied to individual viewpoints. Yet, multi-view strategies demand precise sensor calibration and view alignment for accurate data integration and coherent segmentation outcomes.

Lawin et al. [223] were the first to project a 3D point cloud from several virtual camera views onto 2D planes. Then, using synthetic images, a multi-stream fully connected network is utilized to predict pixel-wise scores. The final semantic label of each point is calculated by combining the re-projected scores from various perspectives. Similarly, Boulch et al. [229] used numerous camera angles to obtain various RGB and depth pictures of a point cloud. They next used 2D segmentation networks to do pixel-by-pixel tagging on these samples. The residual correction [230] is used to merge the scores predicted from RGB and depth pictures.

In order to address the information loss issue, SnapNet [231] takes some selected snapshots of the point cloud to generate pairs of RGB and depth images. They then categorize each pair of 2D photos pixel by pixel using a fully convolutional network. Finally, to complete the work, this model projects the marked points into 3D space. SnapNet attempts to solve the problem of information loss, but it runs into issues throughout the image production process. Consequently, SnapNet-R2 [232] is proposed as a solution for SnapNet. It directly analyses multiple views to produce dense 3D point markers, which improves the segmentation result. The process of creating a labeled point cloud can be broken down into two parts: the labeling of SnapNet 3D and the 2D labeling of RGB-D images extracted from stereo images. Although the model provides a technique that makes it simple to implement, its segmentation accuracy on object boundaries still needs to be improved.

Viewpoint selection and occlusions impact multi-view segmentation algorithms, but they also suffer from information loss and blurring effects due to many-to-one mapping. The nearest predicted label (NLA) strategy improves occluded location processing over K-nearest neighbor (KNN). Processing point cloud data is computationally expensive, and existing projection-based methods have accuracy or parameter issues. The Multi-scale Interaction Network (MINet) [233] balances resources across scales, enhancing efficiency and outperforming point-based, image-based, and projection-based techniques in accuracy, parameter count, and runtime.

### 5.1.2 Range-view based methods

Range view methods for point cloud segmentation processing leverage the inherent spatial structure of the data to preserve fine-grained geometric details, akin to human perception. Processing the point cloud directly in its original form enhances accuracy for tasks requiring precise distance and angle measurements. While these methods minimize preprocessing, they maintain local geometric context due to point proximity, aiding analysis and classification. Range view techniques also align well with diverse sensors and capture devices, facilitating integration into real-world applications. Yet, these methods may be sensitive to sensor viewpoint changes, possibly introducing data inconsistencies and compromising robustness in processing and interpretation.

Wu et al. [224] developed an end-to-end network based on SqueezeNet [234] and Conditional Random Fields (CRF) to perform quick and accurate segmentation of 3D point clouds. Later, they came up with another version named SqueezeSegV2 [235], a segmentation pipeline that uses an unsupervised domain adaption pipeline to solve domain shift and increase segmentation accuracy. To process LiDAR images, all of these methods use conventional convolutions, which is problematic since convolution filters pick up local features that are only active in specific portions of the image. As a result, the network's capacity is underutilized, and segmentation performance suffers. To address this, the author presented SqueezeSegV3 [236], an updated version of the previous SqueezeSeg [224] models that uses Spatially-Adaptive Convolution (SAC) to apply various filters to different regions depending on the input image.

RangeNet++ by Milioto et al. [237] enables real-time semantic segmentation of LiDAR point clouds. It employs GPU-enabled KNN-based postprocessing to address discretization errors and blurry inference outputs after converting 2D range image labels to 3D point clouds. Spherical projection preserves more information than single-view projection, but it may introduce issues like discretization mistakes and occlusions. Lite-HDSeg [238] is another real-time 3D LiDAR point cloud segmentation method. It utilizes a new encoder-decoder architecture with light-weight harmonic dense convolutions. Additionally, the authors introduce ICM, an improved global contextual module capturing multi-scale contextual data, and MCSPN, a multi-class Spatial Propagation Network refining semantic boundaries substantially.

Zhao et al. proposed a projection-based LiDAR semantic segmentation pipeline with a unique network topology and efficient postprocessing [239]. Their FIDNet incorporates a parameter-free FID module that directly upsamples multi-resolution feature maps using bilinear interpolation. It improves model complexity while preserving performance. In contrast to previous methods, this approach maintains neighborhood information more effectively and considers temporal information in single scan segmentation tasks. To address these issues, Wang et al. [240] presented Meta-RangeSeg, which adopts a unique range residual image representation to collect spatial-temporal information. To capture the meta features, Meta-Kernel is used, which minimizes the discrepancy between the 2D range image coordinates input and the Cartesian coordinates output. The multi-scale features were extracted using an efficient U-Net backbone. Moreover, the Feature Aggregation Module (FAM) gathers meta features and multi-scale features, enhancing the range channel's role.

GFNet [241] is based on a Geometric Flow Network (GFN), which can learn the geometric relationships between different views of a 3D point cloud. The GFN comprises a feature extractor and a geometric flow network. The feature extractor captures features from the 3D point cloud, while the geometric flow network learns geometric relationships across different views. These relationships facilitate fusion of features, leading to enhanced semantic segmentation accuracy. GFNet has several advantages over traditional methods for semantic segmentation of 3D point clouds. It can accommodate the irregular and unstructured nature of 3D point clouds by utilizing a deep learning model capable of learning from non-grid data.

CeNet [242] is an efficient method for semantic segmentation of LiDAR point clouds. It utilizes a compact CNN architecture, resulting in faster training and inference due to a reduced parameter count. CeNet consists of three main components: a feature extractor to capture point cloud features, a spatial attention module for emphasizing important features, and a temporal attention module for integrating features across LiDAR sequence frames. In [243], a novel range view representation for LiDAR point clouds is introduced. Based on a CNN architecture, RangeFormer extracts features from the range view representation. These features are utilized by the spatial attention module to learn spatial relationships between points, while the temporal attention module fuses features from different frames. Finally, a decoder predicts the semantic label for each point in the range view.

LENet [244] is a compact and resource-efficient network for LiDAR point cloud semantic segmentation. It incorporates a novel multi-scale convolution attention module that captures long-range dependencies. By utilizing convolutions with varying kernel sizes, features are extracted at multiple scales. Attention mechanisms are employed to assign weights to features from different scales, improving the network's precision in learning semantic segmentation predictions.

### 5.1.3 Bird's eye view-based methods

Bird's-Eye View (BEV) is a 2D representation obtained by projecting a 3D point cloud onto a top-down plane. It provides a flattened view of the point cloud, enabling the application of 2D image-based segmentation techniques. BEV is widely used in point cloud segmentation for tasks like object detection and road segmentation in autonomous driving and robotics. It facilitates the analysis of spatial relationships and captures valuable geometric and contextual information in the horizontal plane.

PolarNet [245] introduces a nearest-neighbor-free segmentation approach for LiDAR data. By converting the Cartesian point cloud to a polar bird's-eye representation, it balances points among grid cells in a polar coordinate system. Polar convolution layers in a deep neural network architecture are utilized to extract features and perform semantic segmentation.

SalsaNet [246] presents an efficient and accurate method for road and vehicle segmentation in LiDAR point clouds for autonomous driving. Its lightweight network architecture incorporates spatial and channel-wise attention mechanisms to capture local and global contextual information. The approach employs a two-step segmentation strategy, using a novel focal loss function to handle class imbalance and improve performance on rare classes.

DGPolarNet [247] addresses the challenges of capturing long-range dependencies and modeling local context by employing a dynamic graph convolutional network. This network dynamically constructs a graph structure based on the input point cloud, capturing spatial relationships between points. Multi-scale features and graph convolutions are utilized to extract discriminative features at different abstraction levels.

Table 5 provides a comprehensive overview of non-point-based methods for semantic segmentation outcomes in 3D point clouds across diverse datasets. In the category of projection-based methods, Rangeformer [243] and MINet [233] achieved the highest results on the nuScenes and SemanticPOSS datasets. However, among models across different methodologies, RangeFormer and DeePr3SS [223] demonstrated superior performance in SemanticKITTI and Semantic3D (red.) datasets.

### 5.2 Discretization-based methods

Discretization-based methods transform continuous point cloud data into a discrete representation for efficient analysis while retaining geometric features. Such discretization serves as a bridge between the raw point cloud and conventional convolutional operations. These methods can be further categorized into two main groups: dense discretization and sparse discretization. Dense discretization involves subdividing the point cloud space into a regular grid and assigning points to corresponding grid cells. This enables the application of standard 3D convolutions, similar to volumetric data. On the other hand, Sparse discretization targets occupied cells, optimizing resource efficiency in line with point cloud sparsity.

#### 5.2.1 Dense discretization representation

Dense Discretization Representation (DDR) converts continuous point clouds into a structured and discrete form using small voxels or grids. This structured representation enables the use of standard 3D convolutional operations and simplifies the handling of irregular and unstructured data. However, it involves a trade-off between resolution, efficiency, and possible discretization artifacts or information loss.

Tchapmi et al. [248] proposed SEGCloud as a means of achieving fine-grained, globally consistent semantic segmentation. Different degrees of geometric relations are first hierarchically abstracted from point clouds in the Fully-Convolutional Point Network (FCPN) [249], and then 3D convolutions and weighted average pooling are used to extract features and incorporate long-range dependencies. This approach can handle large-scale point clouds and has strong inference scalability.

ScanComplete [250] proposed a method for 3D scan completion and per-voxel semantic tagging. It utilizes fully-convolutional neural networks that can adapt to different input data sizes during training and testing. A coarse-to-fine approach is employed to enhance the resolution of predicted results. The volumetric representation preserves the neighborhood structure of 3D point clouds and allows for direct use of 3D convolutions. These factors contribute to the improved performance in this field. However, the voxelization stage introduces discretization artifacts and information loss.

In Cylinder3D [251], a comprehensive analysis of various representations and backbones in 2D and 3D spaces is carried out to determine the usefulness of 3D representations and networks in LiDAR segmentation. It proposes a 3D cylinder partition and convolution-based framework to leverage the 3D topology relations and structures of driving-scene point clouds. Additionally, a context modeling module based on dimension decomposition is introduced to capture high-rank context information progressively.

#### 5.2.2 Sparse discretization representation

Sparse Discretization Representation (SDR) selects a subset of points from point clouds for analysis, offering memory efficiency and computational speed advantages. However, it may struggle with preserving fine details and dense spatial relationships. SDR techniques address these challenges through adaptive sampling and contextual information incorporation.

Choy et al. [252] proposed the MinkowskiNet, a 4D spatio-temporal convolutional neural network for 3D video perception. To properly process high-dimensional data, a generalized sparse convolution is presented. To ensure consistency, a trilateral-stationary conditional random field is used. To encode the local geometrical structures within each voxel, Meng et al. [226] proposed a kernel-based interpolated variational autoencoder architecture. To generate a continuous representation and capture the distribution of points in each voxel, RBFs are used for each voxel instead of the binary occupancy representation. A VAE is also used to map each voxel's point distribution to a compact latent space. Then, to achieve robust feature learning, both symmetry groups and an equivalence CNN are used. Volumetric based networks can be trained and evaluated on point clouds of various spatial sizes due to the scalability of 3D CNN.

Furthermore, Rosu et al. [253] introduced LatticeNet as a way to analyze large point clouds efficiently. DeformsSlice, a data-dependent interpolation module, is also included to back project the lattice feature to point clouds. SPVConv by Tang et al. [254] introduces a lightweight 3D module that combines a high-resolution point-based branch with Sparse Convolution. This module efficiently preserves fine details in large outdoor landscapes. The authors further explore efficient 3D models using SPVConv and conduct a 3D Neural Architecture Search (3D-NAS) to discover optimal network architectures for improved performance across a diverse design space.

SVASeg [255] utilizes Sparse Voxel-based Attention (SVHA) to capture long-range dependencies between sparse points in point clouds. SVHA module points into local regions, computes attention weights, and aggregates features from neighboring regions to predict semantic labels. This approach offers advantages in learning long-range dependencies and is efficient, allowing training and inference on a single GPU. Swin3D [256] is a pretrained transformer backbone for 3D indoor scene understanding. It is based on the Swin Transformer [257] architecture capable of capturing long-range dependencies among points within a 3D point cloud. Swin3D performs self-attention on sparse voxels with linear memory complexity and effectively captures the irregular nature of point signals through generalized contextual relative positional embedding.

In Table 5, the results reveal that among discretization-based methods, Cylinder3D [258], SVASeg [255], and MS1_DVS [259] emerge as the top performers in SemanticKITTI, nuScenes, and Semantic3D (reduced) datasets. However, across various methodologies, MS1_DVS [259] and Swin3D-L [256] excel, surpassing all other approaches in the Semantic3D SanNet, S3DIs (area-5 and 6-fold) datasets.

### 5.3 Hybrid methods

DRINet++ [260] leverages the voxel-as-point concept to enhance the geometric and sparse characteristics of point clouds. It consists of two key modules: Sparse Feature Encoder and Sparse Geometry Feature Enhancement, designed for efficiency and performance improvement. The Sparse Geometry Feature Enhancement improves geometric attributes through multi-scale sparse projection and fusion, while the Sparse Feature Encoder captures local context information. PIG-Net [261], proposed by Hedge et al., adopts a point-inception-based deep neural network for 3D point cloud segmentation. By incorporating an inception module-based inception layer, PIG-Net effectively extracts local features, leading to enhanced performance. To prevent overfitting, Global Average Pooling (GAP) is employed as a regularization technique.

To address the challenge of limited data availability Yan et al. in [262] proposed JS3C-Net that leverages contextual shape priors learned from scene completion and then uses these priors to improve the segmentation of sparse point clouds. JS3C-Net consists of two main components: a scene completion network and a segmentation network. The scene completion network is responsible for predicting a dense point cloud from a sparse point cloud. The segmentation network is responsible for predicting the semantic labels of the dense point cloud.

(AF)2-S3Net [72] employs attentive feature fusion with adaptive feature selection to enhance the segmentation accuracy of sparse point clouds. Comprising a feature extractor, attentive feature fusion module, and segmentation network, it extracts features from the point cloud, fuses them attentively, and predicts semantic labels based on adaptive feature selection using an attention mechanism.

RPVNet [274] is an efficient range-point-voxel fusion network for LiDAR point cloud segmentation. It consists of three branches: range, point, and voxel, which extract features from the range image, point cloud, and voxelized point cloud, respectively. These features are then intelligently fused using a Gated Fusion Module (GFM) to achieve state-of-the-art performance. The GFM selectively combines the relevant features from the three branches for each point.

Hou et al. [277] proposed Point-to-Voxel Knowledge Distillation (PVD), a hybrid method for semantic segmentation of LiDAR point clouds. PVD utilizes knowledge distillation by training a large teacher network on a large dataset and using its point-level predictions to train a small student network. The student network learns from the point-level predictions of the teacher network to achieve accurate semantic segmentation. 2DPASS [278] proposed another hybrid method that combines 2D and 3D information for LiDAR point cloud semantic segmentation. It extracts features from both a 2D image and a 3D grid, fusing them to generate point-

Table 5 Comparative 3D point cloud semantic segmentation result on various available datasets

| Model Name | Year | Semantic KITTI (mIoU) | Semantic POSS (mIoU) | nuScenes (mIoU) | ScanNet (OA) | ScanNet (mIoU) | Semantic3D (OA) | Semantic3D(red.) (mIoU) | S3DIS(Area5) (OA) | S3DIS(Area5) (mIoU) | S3DIS(6-fold) (OA) | S3DIS(6-fold) (mIoU) |
|---|---|---|---|---|---|---|---|---|---|---|---|---|
| **Projection Based Methods** | | | | | | | | | | | | |
| SnapNet [231] | 2017 | – | – | | – | – | 88.60% | 59.10% | – | – | – | – |
| DeePr3SS [223] | 2017 | – | – | | – | – | **88.90%** | 58.50% | – | – | – | – |
| SqueezeSeg [224] | 2018 | 29.50% | 16.80% | | – | – | | | – | – | – | – |
| SqueezeSeg+CRF [224] | 2018 | 30.80% | 18.70% | | – | – | | | – | – | – | – |
| SqueezeSegV2 [235] | 2019 | 39.70% | 29.80% | | – | – | | | – | – | – | – |
| SqueezeSegV2+ CRF [235] | 2019 | 39.60% | 28.90% | | – | – | | | – | – | – | – |
| DarkNet21Seg [38] | 2019 | 47.40% | – | | – | – | | | – | – | – | – |
| DarkNet53Seg [38] | 2019 | 49.90% | – | | – | – | | | – | – | – | – |
| RangeNet53 [237] | 2019 | 49.90% | 25.40% | | – | – | | | – | – | – | – |
| RangeNet21 [237] | 2019 | 47.40% | – | | – | – | | | – | – | – | – |
| RangeNet53++ [237] | 2019 | 52.20% | 28.90% | 65.50% | – | – | | | – | – | – | – |
| KPRNet [263] | 2020 | 63.10% | – | | – | – | | | – | – | – | – |
| PolarNet [245] | 2020 | 54.30% | – | 69.40% | – | – | | | – | – | – | – |
| 3D-MiniNet-KNN [264] | 2020 | 55.80% | – | | – | – | | | – | – | – | – |
| SqueezeSegV3-21 [236] | 2020 | 51.60% | – | | – | – | | | – | – | – | – |
| SqueezeSegV3-53 [236] | 2020 | 55.90% | – | | – | – | | | – | – | – | – |
| SalsaNet [246] | 2020 | 45.40% | – | | – | – | | | – | – | – | – |
| SalsaNext [265] | 2020 | 59.50% | – | 72.20% | – | – | | | – | – | – | – |
| DeepTemporalSeg [266] | 2020 | 37.60% | – | | – | – | | | – | – | – | – |
| AMVNet [267] | 2020 | 65.30% | – | 76.10% | – | – | | | – | – | – | – |
| MPF [268] | 2021 | 55.50% | – | | – | – | | | – | – | – | – |
| TORNADONet [269] | 2021 | 61.10% | – | | – | – | | | – | – | – | – |
| TORNADONet-HiRes [269] | 2021 | 63.10% | – | | – | – | | | – | – | – | – |
| Lite-HDSeg [238] | 2021 | 63.80% | – | | – | – | | | – | – | – | – |
| MINet [233] | 2021 | 52.40% | 30.50% | | – | – | | | – | – | – | – |
| MINet+k-NN [233] | 2021 | 55.20% | **35.10%** | | – | – | | | – | – | – | – |
| FPS-Net [270] | 2021 | 57.10% | – | | – | – | | | – | – | – | – |
| FIDNet [239] | 2021 | 59.50% | – | | – | – | | | – | – | – | – |
| Meta-RangeSeg [240] | 2022 | 61.00% | – | | – | – | | | – | – | – | – |
| DGPolarNet [247] | 2022 | 56.50% | – | | – | – | | | – | – | – | – |
| GFNet [241] | 2022 | 65.40% | – | 76.10% | – | – | | | – | – | – | – |
| CENet [242] | 2022 | 64.70% | – | 74.70% | – | – | | | – | – | – | – |
| RangeFormer [243] | 2023 | **73.30%** | – | **80.10%** | – | – | | | – | – | – | – |
| LENet [244] | 2023 | 64.20% | – | | – | – | | | – | – | – | – |

**Table 5** continued

| Model Name | Year | Semantic KITTI (mIoU) | Semantic POSS (mIoU) | nuScenes (mIoU) | ScanNet (OA) | ScanNet (mIoU) | Semantic3D (OA) | Semantic3D(red.) (mIoU) | S3DIS(Area5) (OA) | S3DIS(Area5) (mIoU) | S3DIS(6-fold) (OA) | S3DIS(6-fold) (mIoU) |
|---|---|---|---|---|---|---|---|---|---|---|---|---|
| **Discretization-based Methods** | | | | | | | | | | | | |
| SEGCloud [248] | 2017 | – | – | – | – | – | 88.10% | 61.30% | 57.35% | 48.90% | – | – |
| SparseConvNet [225] | 2018 | – | – | – | – | 72.50% | – | – | – | – | – | – |
| MS1_DVS [259] | 2018 | – | – | – | – | – | 84.80% | 57.10% | – | – | – | – |
| MS3_DVS [259] | 2018 | – | – | – | – | – | 88.40% | **65.30%** | 57.93% | 46.32% | – | – |
| SPLATNet [271] | 2018 | 18.40% | – | – | – | 39.30% | – | – | – | – | – | – |
| MinkowskiNet [252] | 2019 | – | – | – | – | 73.60% | – | – | 71.71% | 65.40% | – | – |
| VV-Net [226] | 2019 | 52.90% | – | – | – | – | – | – | – | – | 87.78% | 78.22% |
| LatticeNet [253] | 2020 | **65.20%** | – | – | – | 64.00% | – | – | – | – | – | – |
| SVASeg [255] | 2022 | 61.80% | – | 74.70% | – | – | – | – | – | – | – | – |
| Cylinder3D [251] | 2021 | – | – | 77.90% | – | – | – | – | – | – | – | – |
| Swin3D-L [256] | 2023 | – | – | – | – | **77.90%** | – | – | – | **74.50%** | – | **79.80%** |
| **Hybrid Methods** | | | | | | | | | | | | |
| 3DMV [227] | 2018 | – | – | – | – | 48.40% | – | – | – | – | – | – |
| PVCNN [272] | 2019 | 39.00% | – | – | – | – | – | – | **86.87%** | 57.63% | – | – |
| UPB [273] | 2019 | – | – | – | – | 63.40% | – | – | – | – | – | – |
| SPVConv [254] | 2020 | 66.40% | – | 77.40% | – | – | – | – | – | – | – | – |
| SPVNAS [254] | 2020 | 66.40% | – | – | – | – | – | – | – | – | – | – |
| JS3C-Net [262] | 2021 | 66.00% | **60.20%** | – | – | – | – | – | – | – | – | – |
| AF2-S3Net [72] | 2021 | 69.70% | – | 62.20% | – | – | – | – | – | – | – | – |
| RPVNet [274] | 2021 | 70.30% | – | – | – | – | – | – | – | – | – | – |
| DRINet++ [260] | 2021 | 70.70% | – | 80.40% | – | – | – | – | – | – | – | – |
| PMF [275] | 2021 | 63.90%* | – | – | **76.90%** | – | – | – | – | – | – | – |
| MVPNet [276] | 2022 | 53.90% | – | – | – | 64.10% | – | – | – | – | – | – |
| DSPoint [79] | 2022 | – | – | – | – | – | – | – | 70.90% | 63.30% | – | – |
| Point Voxel Transformer [81] | 2022 | 64.90% | – | -76.00% | – | – | – | – | – | **68.20%** | **88.30%** | 69.20% |
| PVKD [277] | 2022 | 71.20% | – | – | – | – | – | – | – | – | – | – |
| Sparse EQ-Net [83] | 2022 | – | – | – | – | **74.30%** | – | – | -71.30% | – | 77.50% | – |
| 2DPASS [278] | 2022 | **72.90%** | – | 80.80% | – | – | – | – | – | – | – | – |
| DeepViewAgg [279] | 2022 | – | – | – | – | – | – | – | -67.20% | – | 74.70% | – |
| LidarMultiNet [280] | 2022 | – | – | **81.40%** | – | – | – | – | – | – | – | – |
| SAT [281] | 2023 | – | – | – | – | 74.20% | – | – | – | – | 72.60% | – |

Here 'OA' refers to the overall accuracy and 'mIoU' denotes mean intersection over union. The symbol '–' means the results are unavailable. The methods are arranged in chronological order within their corresponding categories. The top-performing methods in each category have been highlighted in bold, while the method(s) achieving the best overall performance across all categories are underlined level predictions. These predictions are then used to perform semantic segmentation on the LiDAR point cloud.

In [281], Zhou et al. introduced Size-Aware Transformer (SAT) for 3D point cloud semantic segmentation. SAT adapts receptive fields to object sizes, incorporating multi-scale features and enabling each point to select its attentive fields. The model includes the Multi-Granularity Attention (MGA) scheme for efficient feature aggregation and the Re-Attention module for dynamic adjustment of attention scores. SAT addresses challenges through point-voxel cross attention and a shunted strategy based on multi-head self-attention. By tailoring receptive fields based on object sizes, SAT improves object understanding and achieves enhanced performance.

Table 5 indicates JS3C-Net [262] and LidarMultiNet [280] outperformed all other methods by achieving the highest mIoU scores on the SemanticPOSS and nuScenes datasets, while PMF [275] and Point Voxel Transformer [81] excelled in achieving the highest OA scores on the Sannet and S3DIS(6-fold) datasets.

# 6 Learning strategies for point based methods in semantic segmentation

For semantic segmentation, we will adopt a similar framework as shape classification. In this section, we have exclusively examined approaches that utilize raw point clouds as input. These methods can be further categorized into two groups based on the type of learning supervision employed: supervised and unsupervised methods. Figure 10 provides a comprehensive categorization of raw point-based approaches for point cloud semantic segmentation. Additionally, Table 6 offers a detailed comparison of raw point-based methods for point cloud semantic segmentation across various datasets. The methods are organized chronologically within their respective categories. The evaluation of each method's performance is based on metrics such as overall accuracy (OA) and mean intersection over union (mIoU).

## 6.1 Supervised training

Similar to 3D shape classification, supervised learning methods for semantic segmentation can be categorized into seven distinct groups: pointwise MLP, hierarchical-based, convolution-based, RNN-based, graph-based, transformer-based, and other approaches. These categories can be further organized into feedforward and sequential training paradigms based on the underlying model architecture and data processing techniques.

### 6.1.1 Pointwise MLP methods

Because of the high efficiency, these methods mainly use shared MLP as the basic unit in their networks. Point-wise features retrieved using shared MLP are unable to capture the local geometry in point clouds as well as point-to-point relations [3]. Several networks have been developed to capture a broader context for each point and learn richer local structures, including methods based on neighboring feature pooling, attention-based aggregation, and local–global feature concatenation.

Chen et al. [282] presented a Local Spatial Aware (LSA) layer that learns spatial awareness weights based on the spatial layouts and local structures of point clouds in order to better represent the spatial distribution of a point cloud. For large-scale point cloud segmentation, Hu et al. [6] suggested RandLA-Net, an efficient and lightweight network. This network makes use of random point sampling to attain a remarkable level of computation and memory efficiency. To collect and maintain geometric characteristics, a local feature aggregation module is also provided. In order to reduce the number of redundant ConvNet channels, Hu et al. [128] proposed a novel concept. DRNet, which identifies the most significant channels for each class (dissect) in an interpretable manner and dynamically runs channels according to classes in need (reconstruct). This significantly reduces the network's parameter usage, resulting in a lower memory footprint.

Raw point cloud data inevitably contains outliers or noise as it is generated through different reconstruction algorithms using 3D sensors. Though the MLP method has proven to be efficient, it still fails to capture the spatial relations which is a major downside for this method. In order to extract motion data from a series of massive 3D LiDAR point clouds, Wang et al. [283] developed PointMotionNet, a point-based spatiotemporal pyramid architecture. A key element of PointMotionNet is a cutting-edge method for point-based spatiotemporal convolution, which utilizes a time invariant spatial adjacent space to detect point correspondences across time and extracts spatiotemporal properties.

PS2-Net [284] is a locally and globally aware deep learning framework for semantic segmentation on 3D scene-level point clouds. It incorporates local structures through EdgeConv and global context through NetVLAD, enabling effective integration of local structures and global context. PS2-Net is permutation invariant, making it suitable for handling unordered point clouds. To capture contextual shape information, Sahin et al. proposed ODFNet [285], which utilizes local point orientation distributions around a point. Cone volumes divide the spherical neighborhood of a point, and statistics within each volume serve as point features. The ODF neural network employs an ODFBlock with MLP layers to process the orientation distribution function, enabling

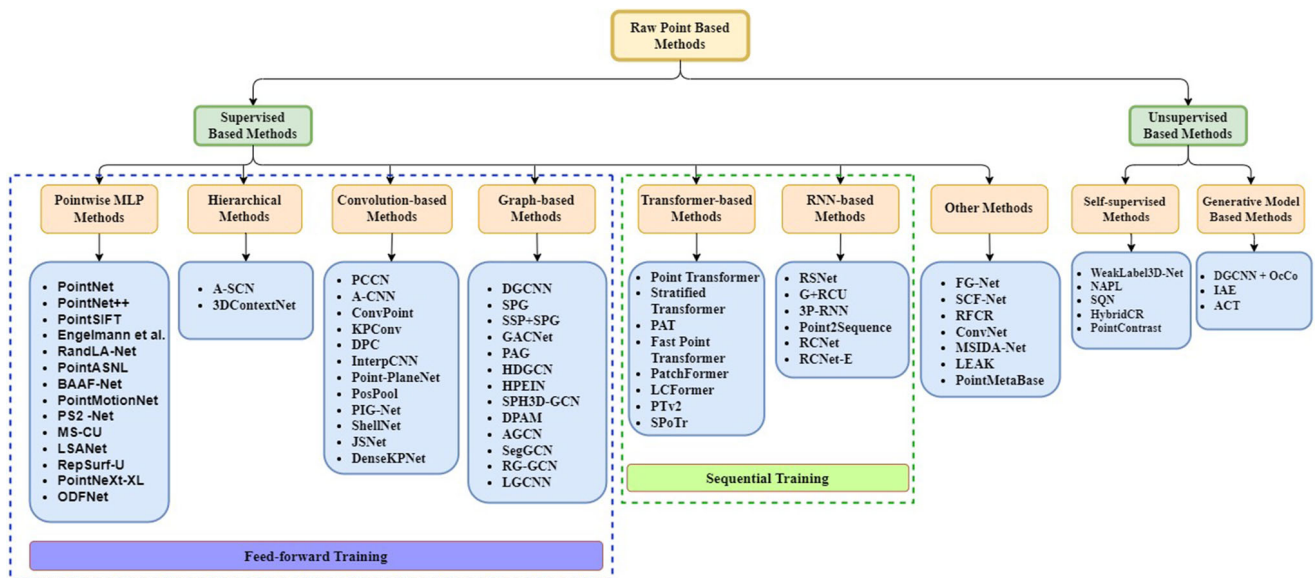

**Fig. 10** A taxonomy of deep learning approaches for raw point-based 3D point cloud semantic segmentation

representation of local patches considering point density distribution along multiple orientations.

Table 6 provides a detailed look at point-based methods for segmenting 3D point clouds in various datasets. Within the pointwise MLP methods category, RandLA-Net [6] stands out with exceptional results in the Semantic3D (red. and sem.) datasets. Meanwhile, PointNeXt-XL [105] and RepSurf-U [286] secure the best Overall Accuracy (OA) in the S3DIS area-5 and 6-fold datasets, respectively. Notably, PointMotionNet [283] achieves the highest mIoU score across all models and methodologies in the SemanticKITTI dataset.

### 6.1.2 Convolution-based methods

Point clouds are a type of 3D data that can be difficult to process with traditional convolution operators. To address this challenge, several approaches have been proposed that use efficient convolution operators specifically designed for point clouds.

The impact of the receptive field on the performance of aggregation-based approaches was demonstrated by Engelmann et al. [287], in their ablation studies with visualization findings. Instead of using the k nearest neighbors, they proposed using a Dilated Point Convolution (DPC) method to aggregate dilated nearby features. This procedure has been shown to be quite successful in boosting receptiveness. Based on kernel point convolution, Thomas et al. [116] suggested a Kernel Point Fully Convolutional Network (KP-FCNN). The euclidean distances to kernel points determine the convolution weights of KPConv, and the number of kernel points is not set. The best coverage places for the kernel points in a sphere space are formulated through an optimization problem. It's important to acknowledge that the radius neighborhood is employed to maintain a consistent receptive field, and grid subsampling is used in each layer to achieve great robustness under variable point cloud densities.

In order to simultaneously handle the instance and semantic segmentation of 3D point clouds, Zhao et al. [288] introduced a novel combined instance and semantic segmentation approach called JSNet. It utilizes an efficient backbone network to extract robust features from the point clouds and a feature fusion module to combine different layer characteristics for more discriminative features. A combined instance semantic segmentation module converts semantic characteristics into instance embedding space and fuses them with instance features. Additionally, it aggregates instance features into a semantic feature space to support semantic segmentation. DPC [287] addresses the issue of limited receptive field size in point convolutional networks for 3D point cloud processing tasks. By expanding the receptive field size of point convolutions, DPC enhances the performance of semantic segmentation and object classification. It can be easily integrated into existing point convolutional networks.

The method proposed in [10], JSENet, is a two-stream network with one stream for semantic segmentation and one stream for edge detection. The two streams share a common encoder, which extracts features from the point cloud. The features from the encoder are then fed into two separate decoders that generate the semantic segmentation and edge detection outputs. In [5], Liu et al. compared different local aggregation operators for point cloud analysis. They explored max pooling, average pooling, and voxel pooling on object classification and segmentation tasks. The study revealed that each operator has unique strengths and weaknesses. Max pooling captures discriminative features but reduces spatial resolution, average pooling finds a balance, and voxel pooling preserves global structures but may lose details.

To address these limitations, the authors introduced a hybrid operator that combines max and average pooling, resulting in improved performance across various tasks. In [289] Li et al. proposed DenseKPNet, which is built on a dense convolutional neural network with Kernel Point Convolution (KPConv). KPConv is a novel convolution operation that is specifically designed for point clouds. It allows the network to learn local features from neighboring points, as well as global features from the entire point cloud.

Table 6 indicates that among pointwise convolution methods, DenseKPNet [289] achieved the highest OA and mIoU in the S3DIS (area-5 and 6-fold) and Semantic3D (red.) datasets. On the other hand, ConvPoint [118] delivered optimal results for Semantic3D (sem.), and KPConv [116] excelled on SemanticKITTI and ScanNet datasets.

### 6.1.3 Hierarchical methods

Hierarchical methods leverage the inherent structural relationships within point clouds to enhance segmentation accuracy and capture finer details. Hierarchical segmentation can lead to improved object part delineation and better handling of complex scenes with varying levels of detail. Given the limited literature on hierarchical data structures in point cloud understanding, and considering the papers covered in the classification section, this discussion focuses on previously unmentioned studies that utilize this approach for semantic segmentation.

Xie et al. [142] developed a novel representation using shape context as a fundamental element in their network architecture. The model can capture and propagate object part information without relying on a fixed grid, and features a simple yet effective contextual modeling mechanism inspired by self-attention based models. Attentional ShapeContextNet (A-SCN) is an end-to-end solution for point cloud classification and segmentation problems.

According to Table 6, 3DContextNet [126] emerged as the top performer within the hierarchical methods category on both the S3DIS datasets.

### 6.1.4 RNN-based methods

Recurrent Neural Networks (RNN) have also been utilized for semantic segmentation of point clouds to capture underlying context information. Bidirectional RNN has been successfully applied to enhance the handling of point clouds in methods like 3P-RNN [290] and RSNets [291], enabling better context capture. RSNets leverage a lightweight local dependence module to capture local structures, incorporating a slice pooling layer, an RNN layer, and a slice unpooling layer. The slice pooling layer projects features from unordered point clouds into an ordered sequence of feature vectors, which are then processed by the RNN layer. 3P-RNN [290] addresses semantic segmentation using raw point clouds, combining a pyramid pooling module and a bidirectional RNN. The pyramid pooling module extracts local spatial data, while the bidirectional RNN captures global context. Inspired by PointNet, 3P-RNN employs pointwise pyramid pooling for local feature acquisition, resulting in faster processing compared to simple pooling in PointNet++.

To convert unordered point feature sets into an ordered series of feature vectors, Huang et al. [291] introduced a lightweight local dependency modeling module that used a slice pooling layer. Addressing the limitations of rigid and fixed pooling operations, Zhao et al. [292] proposed the Dynamic Aggregation Network (DAR-Net), which considers both the global scene intricacy and local geometric factors. DAR-Net employs self-adaptive receptive fields and node weights to dynamically aggregate inter-medium features.

Table 6 demonstrates that within the category of RNN-based methods, 3P-RNN attained the peak OA and mIoU in the S3DIS (area-5) dataset. Simultaneously, RSNet achieved the highest mIoU for the S3DIS (6-fold) dataset.

### 6.1.5 Transformer-based methods

Lai et al. [320] introduced the Stratified Transformer, which effectively captures long-range contexts while maintaining good generalization and performance. The network samples neighboring points within a window as keys for each query point and sparsely samples remote points. This approach allows for a larger receptive field with minimal additional calculations, encompassing both denser local points and sparser distant points. The method incorporates first-layer point embedding to aggregate local information, accelerating convergence and improving performance. Contextual relative position encoding is employed to capture adaptable position information. Additionally, a memory-efficient technique addresses the challenge of fluctuating point counts in each window. Existing point transformers suffer from quadratic complexity in generating attention maps, making them computationally expensive.

Zhang et al. proposed Patchformer [165] which addresses this flaw by combining Patch Attention (PAT) and the Multi-Scale Attention (MST) module to progressively learn a significantly smaller range of bases for computing attention maps. Through a weighted summation on these bases, PAT captures the whole shape context while simultaneously attaining linear complexity for the input size. In the meantime, the model receives multi-scale features from the MST block, which creates attention among features of various scales.

**Table 6** Comperative 3D point cloud semantic segmentation results on various available datasets for Point based methods

| Model Name | Year | SemanticKITTI (mIoU) | S3DIS(Area5) (OA) | S3DIS(Area5) (mIoU) | S3DIS(6-fold) (OA) | S3DIS(6-fold) (mIoU) | ScanNet (OA) | ScanNet (mIoU) | Semantic3D(sem.) (OA) | Semantic3D(sem.) (mIoU) | Semantic3D(red.) (OA) | Semantic3D(red.) (mIoU) | STPLS3D (mIoU) | SensatUrban (mIoU) |
|---|---|---|---|---|---|---|---|---|---|---|---|---|---|---|
| **Pointwise MLP Methods** | | | | | | | | | | | | | | |
| PointNet [3] | 2017 | 14.60% | – | 41.10% | 78.60% | 47.60% | – | 14.69 | – | – | – | – | – | 52.53% |
| PointNet++ [4] | 2017 | 20.10% | – | – | 81.00% | 54.50% | 84.50% | 33.90% | 85.70% | 63.10% | – | – | 15.92% | 58.13% |
| MS-CU [293] | 2017 | – | – | – | 79.20% | 47.80% | – | – | – | – | – | – | – | – |
| PointSIFT [294] | 2018 | – | – | – | 88.70% | 70.20% | 86.20% | 41.50% | – | – | – | – | – | – |
| Engelmann et al. [295] | 2018 | – | 84.20% | 52.20% | 84.00% | 58.30% | 85.10% | – | – | – | – | – | – | – |
| LSANet [282] | 2019 | – | – | – | 86.80% | 62.20% | – | – | – | – | – | – | – | – |
| RandLA-Net [6] | 2020 | 53.90% | – | – | 88.00% | 70.00% | – | 63.00% | 94.60% | 74.80% | 94.80% | 77.40% | 50.53% | 62.80% |
| PointASNL [7] | 2020 | 46.80% | – | 68.70% | – | – | – | – | – | – | – | – | – | – |
| HPGCNN [139] | 2020 | 50.50% | – | – | 90.30% | 69.20% | – | – | – | – | – | – | – | – |
| PS2-Net [284] | 2020 | – | 84.60% | 52.95% | 88.22% | 66.60% | 87.21% | 44.90% | – | – | – | – | – | – |
| BAAF-Net [284] | 2021 | 59.90% | – | 65.40% | 88.90% | 72.20% | – | – | – | – | 94.90% | 75.40% | – | – |
| RPNet-D27 [103] | 2021 | – | – | – | 70.80% | – | 68.20% | – | – | – | – | – | – | – |
| ASSANet [140] | 2021 | – | – | 66.80% | – | – | – | – | – | – | – | – | – | – |
| PointMotionNet [283] | 2022 | 81.80% | – | – | – | – | – | 70.00% | – | – | – | – | – | – |
| RepSurf-U [286] | 2022 | – | 90.20% | 68.90% | 90.80% | 74.30% | – | – | – | – | – | – | – | – |
| PointNeXt-XL [105] | 2022 | – | 90.60% | 70.50% | 90.30% | 74.90% | – | 71.20% | – | – | – | – | – | – |
| ODFNet [285] | 2022 | – | – | – | 90.80% | 72.20% | – | – | – | – | – | – | – | – |
| **Convolution-based Methods** | | | | | | | | | | | | | | |
| PCCN [296] | 2018 | – | – | 58.30% | – | – | – | – | – | – | – | – | – | – |
| TangentConv [297] | 2018 | 35.90% | 82.50% | 52.80% | – | – | 80.10% | 40.90% | – | – | – | – | – | – |
| PointCNN [145] | 2018 | – | 85.91% | 66.36% | 88.14% | 75.61% | 79.70% | 55.70% | – | – | – | – | – | – |
| A-CNN [148] | 2019 | – | – | – | 87.30% | – | 85.40% | – | – | – | – | – | – | – |
| KPConv [116] | 2019 | 58.80% | – | 67.10% | 68.50% | 67.10% | – | 68.40% | – | – | 92.90% | 74.60% | 53.73% | 64.50% |
| InterpCNN [109] | 2019 | – | – | – | 88.70% | 66.70% | – | – | – | – | – | – | – | – |
| ShellNet [111] | 2019 | – | – | – | 67.90% | 66.80% | 85.20% | – | – | – | 93.20% | 69.40% | – | – |
| PointConv [119] | 2019 | – | – | – | 88.70% | 61.70% | – | – | – | – | – | – | – | – |
| JSNet [288] | 2019 | – | 87.70% | 54.50% | – | – | 69.90% | – | – | – | – | – | – | – |
| JSENet [10] | 2020 | – | – | 67.70% | – | – | – | – | – | – | – | – | – | – |
| PosPool [5] | 2020 | – | – | 66.70% | – | 54.80% | – | – | – | – | – | – | – | – |
| Point-PlaneNet [112] | 2020 | – | – | – | 83.90% | – | – | – | – | – | – | – | – | – |
| ConvPoint [118] | 2020 | – | 88.80% | – | 88.80% | 68.20% | – | – | 93.40% | 76.50% | – | – | – | – |
| DPC [287] | 2020 | – | 86.80% | 61.30% | – | – | – | 59.20% | – | – | – | – | – | – |
| DenseKPNet [289] | 2022 | – | 90.80% | 68.90% | 89.30% | 71.90% | – | – | – | – | 94.90% | 77.90% | – | – |

Table 6 continued

| Model Name | Year | SemanticKITTI (mIoU) | S3DIS(Area5) (OA) | S3DIS(Area5) (mIoU) | S3DIS(6-fold) (OA) | S3DIS(6-fold) (mIoU) | ScanNet (OA) | ScanNet (mIoU) | Semantic3D(sem.) (OA) | Semantic3D(sem.) (mIoU) | Semantic3D(red.) (OA) | Semantic3D(red.) (mIoU) | STPLS3D (mIoU) | SensatUrban (mIoU) |
|---|---|---|---|---|---|---|---|---|---|---|---|---|---|---|
| **Hierarchical Methods** | | | | | | | | | | | | | | |
| A-SCN [142] | 2018 | – | – | – | 81.59% | 52.72% | – | – | – | – | – | – | – | – |
| 3DContextNet [126] | 2018 | – | **84.90%** | **55.60%** | **90.60%** | **72.00%** | – | – | – | – | – | – | – | – |
| *RNN-based Methods* | | | | | | | | | | | | | | |
| G+RCU [293] | 2017 | – | – | 45.10% | 81.10% | 49.70% | – | – | – | – | – | – | – | – |
| RSNet [291] | 2018 | – | – | 51.93% | – | **56.50%** | – | 39.35% | – | – | – | – | – | – |
| 3P-RNN [290] | 2018 | – | **86.90%** | **56.30%** | **85.70%** | 53.40% | – | – | – | – | – | – | – | – |
| RCNet [157] | 2019 | – | – | – | 82.01% | 51.40% | – | – | – | – | – | – | – | – |
| RCNet-E [157] | 2019 | – | – | – | 83.58% | 53.21% | – | – | – | – | – | – | – | – |
| **Graph-based Methods** | | | | | | | | | | | | | | |
| SPG [298] | 2018 | – | 86.38% | 58.04% | 85.50% | 62.10% | – | – | **92.90%** | **76.20%** | **94.00%** | **73.20%** | – | – |
| DGCNN [132] | 2018 | – | – | 47.10% | 84.10% | 56.10% | – | – | – | – | – | – | – | – |
| SSP+SPG [299] | 2019 | – | 87.90% | 61.70% | 87.90% | 68.40% | – | – | – | – | – | – | – | – |
| PointWeb [131] | 2019 | – | 87.00% | 60.30% | 87.30% | 66.70% | 85.90% | – | – | – | 91.90% | 70.80% | – | – |
| GACNet [300] | 2019 | – | 87.79% | 62.85% | – | – | – | – | – | – | – | – | – | – |
| HDGCN [301] | 2019 | – | – | 59.33% | – | 66.85% | – | – | – | – | – | – | – | – |
| Jiang et al. [302] | 2019 | – | 87.18% | 61.85% | 88.20% | 67.83% | – | **61.80%** | – | – | – | – | – | – |
| DPAM [134] | 2019 | – | 86.10% | 60.00% | 87.60% | 64.50% | – | – | – | – | – | – | – | – |
| PAG [303] | 2020 | – | **88.81%** | **69.32%** | 88.10% | 65.90% | – | – | – | – | – | – | – | – |
| SegGCN [304] | 2020 | – | 88.20% | 63.60% | – | – | – | 58.90% | – | – | – | – | – | – |
| SPH3D-GCN [305] | 2021 | – | 87.70% | 59.50% | **88.60%** | **68.90%** | – | 61.00% | – | – | – | – | – | – |
| RG-GCN [306] | 2022 | – | – | – | 88.10% | 63.70% | – | – | – | – | – | – | – | – |

**Table 6** continued

| Model Name | Year | SemanticKITTI (mIoU) | S3DIS(Area5) (OA) | S3DIS(Area5) (mIoU) | S3DIS(6-fold) (OA) | S3DIS(6-fold) (mIoU) | ScanNet (OA) | ScanNet (mIoU) | Semantic3D(sem.) (OA) | Semantic3D(sem.) (mIoU) | Semantic3D(red.) (OA) | Semantic3D(red.) (mIoU) | STPLS3D (mIoU) | SensatUrban (mIoU) |
|---|---|---|---|---|---|---|---|---|---|---|---|---|---|---|
| **Transformer-based Methods** | | | | | | | | | | | | | | |
| PAT [159] | 2019 | – | – | 60.10% | – | 64.30% | – | – | – | – | – | – | – | – |
| FastPointTransformer [307] | 2021 | – | – | 70.10% | – | – | – | 72.10% | – | – | – | – | – | – |
| PointTransformer [160] | 2021 | – | 90.80% | 70.40% | 90.20% | 73.50% | – | – | – | – | – | – | 47.64% | – |
| Spg | 2022 | – | 91.50% | 72.00% | – | – | – | 73.70% | – | – | – | – | – | – |
| PatchFormer [165] | 2022 | – | – | 68.10% | – | – | – | – | – | – | – | – | – | – |
| LCPFormer [167] | 2022 | – | 90.80% | 70.20% | – | – | – | – | – | – | – | – | – | 63.40% |
| PTv2 [166] | 2022 | – | – | 71.60% | – | – | – | 75.20% | – | – | – | – | – | – |
| StratifiedFormer+PAGWN [308] | 2022 | – | 91.40% | 72.20% | 91.70% | 77.60% | – | – | – | – | – | – | – | – |
| SPoTr [168] | 2023 | – | 90.70% | 70.80% | – | – | – | – | – | – | – | – | – | – |
| **Unsupervised-based Methods** | | | | | | | | | | | | | | |
| PointConstrast [309] | 2020 | – | – | 70.90% | – | – | – | – | – | – | – | – | – | – |
| DGCNN-OcCo [191] | 2021 | – | – | – | 84.60% | 58.00% | – | – | – | – | – | – | – | – |
| GuidedPoint [310] | 2021 | 67.70% | – | 68.80% | – | – | – | 74.00% (V2) | – | – | – | – | – | – |
| IAE [193] | 2022 | – | 85.90% | 60.70% | – | – | – | – | – | – | – | – | – | – |
| ACT [203] | 2022 | – | 71.10*% | 61.20% | – | – | – | – | – | – | – | – | – | – |
| Hao et al. [201] | 2022 | – | 90.10% | 69.30% | 90.20% | 73.50% | – | 75.80% | – | – | – | – | – | – |
| HybridCR [311] | 2022 | 52.30% | – | 51.50% | – | 69.20% | – | 56.80% (v2) | – | – | – | 76.80% | – | – |
| NAPL [312] | 2022 | 61.60% | – | – | – | – | – | – | – | – | – | – | – | – |
| SQN [313] | 2022 | 50.80% | – | 61.40% | – | – | – | 56.90% | 94.80% | 72.30% | 93.70% | 74.70% | – | 54.00% |
| WeakLabel-3DNet [314] | 2022 | 53.70% | – | 68.10% | – | – | – | 67.90% | – | 75.30% | – | – | – | – |
| **Other Methods** | | | | | | | | | | | | | | |
| FG-Net [173] | 2021 | 53.80% | 88.20% | – | – | 70.80% | – | 68.50% | 93.60% | 78.20% | – | – | – | – |
| SCF-Net [315] | 2021 | – | 88.40% | 71.60% | – | – | – | – | – | – | 94.70% | 77.60% | 50.65% | – |
| KPConv+RFCR [316] | 2021 | – | – | 68.73% | – | 71.70% | – | 70.20% | – | – | – | 77.80% | – | – |
| Shao et al. [317] | 2022 | – | – | – | 88.90% | 71.70% | – | – | – | – | 95.30% | 79.00% | – | 63.60% |
| ConvNet+CBL [318] | 2022 | – | 90.60% | 69.40% | 89.60% | 73.10% | – | 70.50% | – | – | 95.00% | 78.40% | – | – |
| MSIDA-Net [319] | 2022 | 59.80% | – | – | 89.20% | 73.00% | – | – | – | – | 94.60% | 77.80% | – | – |
| LEAK(Cylinder3D) [258] | 2023 | 65.20% | – | – | – | – | – | – | – | – | – | – | – | – |
| LEAK(RandLA-Net) [258] | 2023 | – | – | – | 90.90% | 76.10% | – | – | – | – | – | – | – | – |
| PointMetaBase-XXL [179] | 2023 | – | 90.80% | 71.30% | 91.30% | 77.00% | – | 71.40% | – | – | – | – | – | – |

Here 'OA' refers to the overall accuracy and 'mIoU' denotes mean intersection over union. The symbol '-' means the results are unavailable. The methods are arranged in chronological order within their corresponding categories. The top-performing methods in each category have been highlighted in bold, while the method(s) achieving the best overall performance across all categories are underlined [308] proposes a novel Window Normalization (WN) module for 3D point cloud understanding. WN is a simple yet effective module that can be easily integrated into existing point cloud neural networks. WN works by normalizing the features of each point in a local window to have unit length. This helps to unify the point densities in different parts of the point cloud, which can improve the performance of point cloud neural networks on tasks such as semantic segmentation and object detection.

According to the results presented in Table 6, Stratified-Former+PAGWN [308] emerged as the top performer for both the S3DIS area-5 and 6-fold datasets among all models from various methodologies.

### 6.1.6 Graph-based methods

Graph networks are used in a variety of methods to capture the underlying forms and geometric features of 3D point clouds.

Ma et al. [321] suggested a Point Global Context Reasoning (PointGCR) module that uses an undirected graph representation to capture global contextual information along the channel dimension. PointGCR is a plug-and-play module that can be trained from beginning to end. To boost performance, it may be readily added into an existing segmentation network. Furthermore, numerous recent studies have attempted to perform semantic segmentation of point clouds with less supervision. For semantic segmentation of point clouds, Xu et al. [322] studied many inexact supervision techniques. They also presented a network that can be trained with points that are only partially tagged.

Kang et al. developed PyramNet [323] using Graph Embedding Module (GEM) and Pyramid Attention Network (PAN). GEM transforms the point cloud into a directed acyclic graph, using a covariance matrix for adjacent similarity instead of euclidean distance. The PAN module extracts features with different semantic intensities using convolution kernels of four distinct sizes. Graph Attention Convolution (GAC) was introduced in [300] as a way to learn useful features from a nearby adjacent set selectively. This is accomplished by assigning attention weights to various surrounding points and feature channels depending on their spatial placements and feature differences. GAC is similar to the widely used CRF model in that it may learn to capture discriminative features for segmentation.

For effective graph convolution of 3D point clouds, Lei et al. [305] proposed a spherical kernel. It quantizes local 3D space systematically to capture geometric relationships. The spherical CNN kernel shares weights for similar structures, providing translation-invariance, and supports precise geometric learning through asymmetry. It eliminates edge-dependent filter generation, making it efficient for large point clouds. Vertices represent points, edges connect neighbors, and coarsening is done using farthest point sampling. The primary challenge in learning from point clouds is capturing local structures and relationships.

The capacity of graph convolution to extract local shape information from neighbors is very powerful. Inspired by this, Lian et al. introduce the Hierarchical Depthwise Graph Convolutional Neural Network (HDGCN) [301]. HDGCN employs a memory-efficient depthwise graph convolution, known as DGConv, along with pointwise convolution. DGConv enables local feature extraction and transfer between points and their neighbors while being order-invariant.

Zeng et al. introduced RG-GCN [306], a Random Graph-based Graph Convolution Network for point cloud semantic segmentation. It comprises two main components: a random graph module that constructs a random graph for each point cloud, and a graph convolution network based on a modified PointNet++ architecture, known for its effectiveness in point cloud semantic segmentation.

Table 6, demonstrates that among the graph-based methods, SPG [298] attained the best OA and mIoU for both the Semantic3D datasets. Meanwhile, PAG [303] yielded the optimal results for S3DIS (area-5), and SPH3D-GCN [305] excelled in the S3DIS (6-fold) dataset.

### 6.1.7 Unsupervised training

PointContrast [309] is an unsupervised pre-training method for 3D point cloud understanding. It employs a contrastive learning framework to learn representations from unlabeled point cloud data. This two-stage method extracts local features by grouping points into patches and encodes them into fixed-dimensional representations. The contrastive loss is then used to encourage similarity among representations from the same patch and dissimilarity among representations from different patches. PointContrast highlights the potential of leveraging unlabeled data for effective representation learning in 3D point cloud analysis.

Authors in [311] proposed a novel Hybrid Contrastive Regularization (HybridCR) framework for weakly-supervised point cloud semantic segmentation. HybridCR leverages both point consistency and contrastive regularization with pseudo labeling in an end-to-end manner. Fundamentally, HybridCR explicitly and effectively considers the semantic similarity between local neighboring points and global characteristics of 3D classes. In their work [314], a novel weakly supervised framework called WeakLabel3D-Net is proposed for understanding real-scene LiDAR point clouds. This multi-task framework achieves state-of-the-art results on various LiDAR datasets, even with limited labeled data. WeakLabel3D-Net comprises a point cloud encoder, task-specific decoders, and a weakly supervised loss function. The encoder extracts features, decoders generate predictions, and the loss function trains the network with labeled data.

The framework utilizes a modified PointNet++ encoder, task-specific decoders, and a combination of cross-entropy and consistency losses to encourage consistent predictions for neighboring points.

Zhao et al. [312] proposed Number-Adaptive Prototype Learning (NAPL), a weakly supervised approach that learns from a small number of labeled points. It learns prototypes by clustering unlabeled points and then predicts the class of a point by finding the closest prototype. What sets NAPL apart is its adaptive learning of the number of prototypes for each class. This is achieved using a novel loss function that penalizes the classifier for assigning the same class to nearby points, encouraging the learning of distinct prototypes even for close points of the same class. For semantic segmentation of large-scale 3D point cloud, Hu et al. proposed Semantic Query Network (SQN) [313], a graph-based method that leverages both point consistency and contrastive regularization. Fundamentally, SQN explicitly and effectively considers the semantic similarity between neighboring 3D points, allowing the extremely sparse training signals to be back-propagated to a much wider spatial region, thereby achieving superior performance under weak supervision.

According to the findings in Table 6, amidst models employing unsupervised methodology [201] showcased the most promising results on both the S3DIS area-5 and 6-fold datasets. However, when considering models spanning all methodologies, [201] outperformed the rest on the ScanNet dataset, while SQN [313] achieved the highest Overall Accuracy (OA) for the Semantic3D (sem.) dataset.

### 6.1.8 Other methods

Fan et al. [315] proposed SCF, a learnable module for extracting Spatial Contextual Features (SCF) from large-scale point clouds. SCF consists of three components: a local polar representation block, a dual-distance attentive pooling block, and a global contextual feature block. The module constructs spatial representations invariant to z-axis rotation, learns discriminative local features using neighboring representations, and incorporates global context based on spatial location and neighborhood volume ratio. Gong et al. [316] proposed Receptive Field Component Reasoning (RFCR) for point cloud segmentation, utilizing Target Receptive Field Component Codes (RFCCs) to guide a coarse-to-fine category reasoning approach. The method incorporates a gradual RFCR module that enhances neural network representation by iteratively reasoning about receptive field components, enabling the learning of progressively complex features. Additionally, a feature densification technique employing centrifugal potential is introduced to improve feature selection by separating positive and negative features.

[319] presented Multispatial Information and Dual Adaptive (MSIDA) module for learning point cloud semantic segmentations. MSIDA addresses challenges posed by disordered and unevenly distributed large-scale 3D point clouds. It includes an MSI block to encode spatial information using cylindrical and spherical coordinate systems, and DA blocks for weighted fusion of local features and improved local region understanding. By incorporating spatial information and adaptive feature integration, the MSIDA module enhances point cloud segmentation, enabling better comprehension of complex geometric structures in scenes.

To address the challenge of unsatisfactory segmentation performance on scene boundaries in 3D point cloud data, [318] introduced metrics to quantify this issue and proposes a Contrastive Boundary Learning (CBL) framework. CBL quantifies the issue and enhances feature discrimination across boundaries by contrasting representations using scene contexts at multiple scales. Camuffo et al. [258] introduced LEAK, which clusters classes into macro groups based on mutual prediction errors for point cloud semantic segmentation. LEAK aligns class-conditional prototypical feature representations for fine and coarse classes to regularize the learning process. This prototypical contrastive learning approach improves generalization across domains and reduces forgetting during knowledge distillation from prototypes. Additionally, it incorporates a per-class fairness index to ensure balanced class-wise results.

Table 6 shows that among the models categorized under other methods, PointMetaBase-XXL [179] demonstrated promising results across multiple datasets, including S3DIS (area-5, 6-fold), and the ScanNet dataset. However, when compared to models from all other methodologies, both FG-Net [173] and [317] excelled in both Semantic3D datasets.

## 7 Discussion and future directions

Although extensive research has been conducted on point cloud processing networks for 3D object classification and segmentation, their performance still falls behind when compared to RGB images. This disparity is due to the irregular and sparse nature of the point cloud. As a result, there is still considerable work to be done.

This section aims to highlight potential key research directions and future applications in a comprehensive manner. The following aspects will be explored to provide insight into the possible directions of this field.

### 7.1 Algorithmic advances

Point cloud processing algorithms are critical to efficiently process the vast amounts of data contained in point clouds. Future algorithms may incorporate more advanced deep learning techniques to better handle the complexities of point cloud data. A possible strategy for dealing with raw

Table 7 A summarization and comparison between existing methods for 3D point cloud understanding

| Methods | Strengths | Weaknesses |
| --- | --- | --- |
| Projection-based | By leveraging 2D convolutional architectures to solve a 3D task, projection based techniques eventually make 3D learning more scalable by bridging the gap between 2D and 3D learning | Generates quantization artifacts that may make it difficult to see the data's inherent invariances. Multi-resolution features and geometric features are not exploited properly. Not appropriate for tasks requiring per-point processing. Due to repeated convolution procedures, spatial information of small instances is lost |
| Voxel-based | Voxel-based models are compatible with traditional 3D convolutions, have regular data locality, and can effectively encode coarse-grained features | Voxelization results in the loss of geometrical and spatial resolution information. It is not scaleable since the computational and memory footprints increase cubically with resolution |
| Range view | Range view methods directly process raw point cloud data, preserving intricate geometry and offering an intuitive spatial representation akin to human perception. By working with raw point coordinates, they maintain underlying geometric information, ideal for tasks demanding precise measurements of distances and angles | Range view methods are viewpoint-sensitive, with changes in sensor placement or orientation introducing data variability that can undermine robust processing and interpretation. Sparse data regions, resulting from sensor characteristics, can challenge these methods, potentially impeding accurate analysis |
| Bird's-Eye View | Bird's-eye view offers a top-down perspective, simplifying tasks like object detection by converting the problem from 3D to 2D. It's more stable against sensor pose changes than range view, which is beneficial in scenarios with varying sensor orientation | Bird's-eye view inherently lose some of the fine geometric details present in the range view, limiting precision in geometric analysis. It excels in horizontal plane understanding but struggles with accurate vertical capture, potentially causing ambiguity due to object overlap at varying heights |
| Hybrid | By integrating high-level image semantics to points, the resolution mismatch issue between dense RGB and sparse depth can be resolved | High coupling between image and LiDAR models lowers overall model reliability and raises development costs |
| Point-based | Point-based models directly learn from sparse and unstructured point clouds while maintaining the accuracy of point location | Usually requires a lot of computing power, especially with the large-scale point cloud dataset |
| Supervised based | Supervised learning can produce highly accurate results when trained on a large dataset of labeled point clouds. Once a model is trained, it can be used to process new point clouds quickly and efficiently. This is especially useful in real-time applications where processing speed is critical | The amount of labeled data needed for supervised learning is significant, and it may struggle with new or unseen data. It can also be biased towards the training data, leading to incorrect results and prone to overfitting |

**Table 7** continued

| Methods | Strengths | Weaknesses |
|---|---|---|
| Unsupervised based | Unsupervised learning can be used to identify patterns and structures in point cloud data without the need for labeled data. It helps to cluster similar objects or segments within point clouds and can identify outliers and anomalies in point cloud data, which may be missed by supervised learning methods | Unsupervised learning methods can be more computationally intensive than supervised learning methods, which can be a limitation when dealing with large point cloud datasets. It lacks labeled data guidance, which can potentially result in less accuracy than supervised learning. The evaluation of unsupervised learning models can be challenging due to the absence of a clear objective function to optimize |

point clouds could be distinct from traditional methods. In comparison to CNN, transformer architectures have recently demonstrated promising accuracy on point cloud learning benchmarks. The self-attention operator, which is at the heart of transformer networks, is invariant to the input elements' permutation and cardinality. As a result, the transformer family of models is admirably adapted to point cloud processing. Self-supervised representation learning on point cloud data has proven to be another promising solution. Self-supervised representation learning analyzes how to properly pre-train deep neural networks with unlabeled and raw input data. Instead of creating representations based on human-defined annotations, self-supervised learning learns latent features from unlabeled data. It is commonly accomplished by creating a pretext assignment to pre-train the model before fine-tuning it on subsequent tasks. Self-supervised learning has significantly enhanced computer vision by reducing reliance on labeled data. The increasing volume of papers on various methods published between 2015 and 2023 underscores the growing research interest in unsupervised networks.

In addition, many researchers are currently interested in optimizing neural network training and model compression. Reducing the parameters of a network can speed up training while also allowing deep learning techniques to be used on devices with limited resources. These advanced algorithms may also incorporate more sophisticated data fusion techniques to integrate and be benefited from different point cloud representation. To handle 3D input, previous research has used either voxel-based or point-based NN models. However, both methods are computationally inefficient. With increasing input resolution, voxel-based models' memory use and computation costs grow cubically. Rather than extracting features, up to 80% of the effort in point-based networks is spent shaping the sparse input, which has poor memory localization. Hence, recent research has focused on maximizing the benefits of both strategies while minimizing the drawbacks. Liu et al. presented Point-Voxel CNN (PVCNN) [272], which represents 3D input data as point clouds to take advantage of sparsity and employs voxel-based convolution to produce a contiguous memory access pattern. (DSPoint) extracted local global features by simultaneously operating on voxels and points, combining both local features and global geometric architecture. A combination of projection and raw point-based approaches is being studied by some researchers. In PointView-GCN [80], the network uses multi-level GCNs to record both the geometrical cues of an object and its multiview relations, which hierarchically collect the shape attributes of single-view point clouds. Table 7 summarizes and compares different existing methods for 3D point cloud understanding.

### 7.2 Improved sensor technology

Sensor technology has witnessed notable advancements, with lidar and photogrammetry systems showing great promise. Lidar sensors, renowned for their precise distance measurement and object detection capabilities, have become more accurate, affordable, and accessible to diverse industries. Multispectral and hyperspectral imaging techniques, complementing point cloud data with material and chemical information, offer opportunities in archaeology, geology, and forestry, enabling detailed analysis and conservation efforts.

Although sensors have improved, there is room for further enhancement. Higher accuracy can minimize errors and enhance point cloud quality, leading to more precise modeling and analysis. Fusion of data from multiple sensors provides a comprehensive understanding of environments, improving accuracy and reducing errors. For instance, the combined use of lidar and photogrammetry systems captures both geometric and textural information. Additionally, sensing techniques like multispectral and hyperspectral imaging provide insights into environmental composition and properties, benefiting various applications.

The ongoing evolution of sensor technology will significantly impact point cloud processing. The availability of more accurate, affordable, and capable sensors facilitates the capture of high-resolution point cloud data, empowering researchers and engineers to make informed decisions and develop advanced solutions.

### 7.3 Advancement in datasets

Point cloud datasets play a vital role in various fields, including autonomous vehicles, robotics, virtual reality, and 3D modeling. While advancements have been made in dataset creation, further developments are needed to enhance their quality and usefulness.

Efficiency improvements in collecting and processing point cloud data are key areas for advancement. Integrating data from multiple sensors, such as LiDAR, cameras, or radar, can provide more comprehensive environmental information and enable algorithms to handle complex scenarios. Diverse datasets that encompass a broader range of environments and scenarios would enhance the performance and accuracy of point cloud processing algorithms. Manual annotation of datasets is time-consuming, and developing automated annotation methods would expedite the process and improve accuracy. Additionally, incorporating temporal information, such as capturing data at different times or using motion-capturing sensors, would enable algorithms to track and predict environmental changes.

Advancements in these areas would greatly improve the practicality and applicability of point cloud datasets in various industries. Ongoing development and improvement of point cloud datasets are essential for advancing the field of point cloud processing and facilitating new applications.

### 7.4 Cloud computing

Cloud computing has significantly impacted point cloud processing, enhancing its accessibility and affordability. In the future, advanced cloud-based processing tools are anticipated to incorporate real-time capabilities and distributed computing, enabling efficient handling of large data volumes in real-time. Real-time processing would facilitate quicker decision-making for critical applications like disaster response and autonomous systems. Evolving cloud-based tools, with specialization and advancements, will unlock new applications and use cases for point cloud data.

Advances in technology, algorithms, and applications will shape the future of point cloud processing, fostering innovation and the emergence of novel applications and use cases.

## 8 Conclusions

This paper presents current state-of-the-art methodologies and recent advancements in 3D shape classification and semantic segmentation. Because of the potential for practical applications such as autonomous driving, robot manipulation, and augmented reality, point cloud understanding has recently gained a lot of attention. Specific deep learning frameworks are designed to match point clouds from several scans of the same scene, and generative networks are adapted to enhance the quality of point cloud data in terms of noise and missing points. Deep learning techniques that are correctly adapted have been found to be effective in addressing the unique challenges presented by point cloud data. A detailed taxonomy is presented, accompanied by a performance evaluation of multiple approaches using widely utilized datasets. The benefits and drawbacks of various methodologies, as well as potential research directions, are also highlighted. We believe that our work stands as a confident and impactful addition to the field, providing a valuable resource for researchers and practitioners alike.

**Acknowledgements** This material is based upon work supported by the National Science Foundation under Grant No. OIA-2148788.

**Author Contributions** Sushmita Sarker authored the main manuscript text. Gunner Stone generated the figures (Fig. 6). Prithul Sarker, Gunner Stone, and Ryan Gorman contributed materials, and verified the accuracy of all information. The manuscript was collectively reviewed by all authors.

**Data availability** No datasets were generated or analysed during the current study.

### Declarations

**Conflict of interest** The authors declare no Conflict of interest.

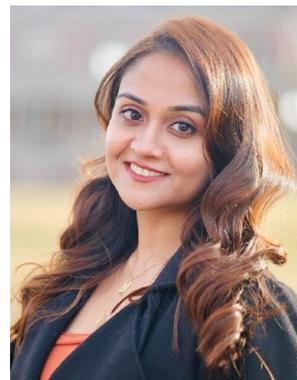

**Sushmita Sarker** is a Ph.D. student in Computer Science and Engineering at the University of Nevada, Reno, USA. She earned her B.Eng. degree in Electronics and Communication Engineering from Gujarat Technological University, India, in 2017 and her Master's degree in Computer Science Engineering from the University of Nevada, Reno, in 2023. Her research interests span computer vision, point cloud processing, and deep learning. Her primary research focuses on the analysis of medical images using generative networks to improve disease prediction and detection. Additionally, she endeavors to utilize point cloud data for mapping sagebrush and understanding the impacts of wildfires on forest ecosystems.

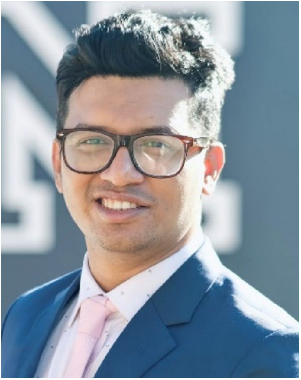

**Prithul Sarker** is a Ph.D. candidate in the Department of Computer Science and Engineering at the University of Nevada, Reno, USA. He received B.S. degree in electrical and electronic engineering from Bangladesh University of Engineering and Technology, Bangladesh in 2017, and M.S. degree in computer science from the University of Nevada, Reno, USA in 2023. His primary research focus is on utilizing virtual reality (VR) technology to assess pupillary function and artificial technology (AI) in healthcare. His research aims to improve the reproducibility and validity of findings related to pupillary function, with a focus on its significance in physiological and psychological processes using VR technologies. Additionally, he has performed comprehensive analyses on a range of medical data modalities, including time series data sourced from assessments and images from medical diagnoses, to predict and detect conditions and diseases such as concussion, breast cancer and ocular disorder.

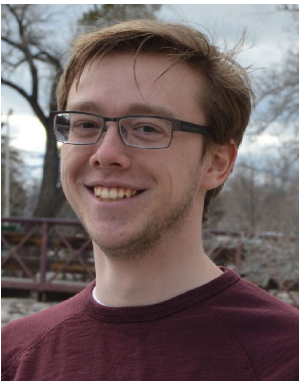

**Gunner Stone** (M.Sc., 2023, B.Sc., 2020) is currently a Ph.D. candidate in Computer Science and Engineering at the University of Nevada, Reno. At the university, he actively collaborates within the Human-Machine Perception Lab and the GEARS Lab. His primary research focus lies in LiDAR pointcloud classification and part segmentation using Deep Learning, especially in the domain of forest ecology. This interest aims to leverage LiDAR pointclouds for mapping forest attributes to understand the impacts of wildfires and climatic changes on forest ecosystems.

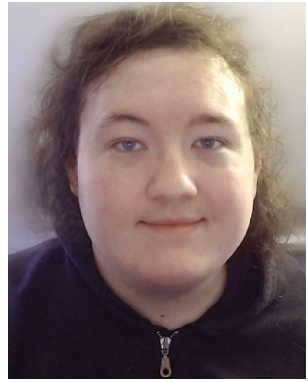

**Ryan C. Gorman** earned his Bachelor of Science degree in Computer Science and Engineering from the University of Nevada, Reno, in 2021. He is currently a Ph.D. student at the same institution in the Department of Computer Science and Engineering. His current research focuses on enhancing wildfire data quality through multi-modality and super-resolution techniques. Ryan's research interests lie in optimizing machine learning and computer vision efficiency while ensuring effectiveness. He is dedicated to making advanced models more accessible, especially for compute limited devices.

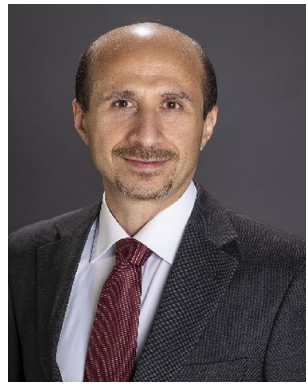

**Javad Sattarvand** is an associate professor and the department chair at the Department of Mining and Metallurgical Engineering at the University of Nevada, Reno, since 2015. Dr. Sattarvand received his Ph.D. in Mining Engineering from the RWTH University of Aachen - Germany in 2009. With 23 years of academic appointments, he has supervised 34 master's and 5 Ph.D. students, received over 4 million dollars in research grants and published over 100 peer-reviewed papers in journals and conferences. He is the teacher for mining automation and mining software engineering courses at the University of Nevada, Reno. With industrial expertise in the optimization of mining operations, mine monitoring, and mine control systems, he has founded three start-up companies and accomplished more than 20 industrial projects and the commercialization of several software and hardware products for mining automation.

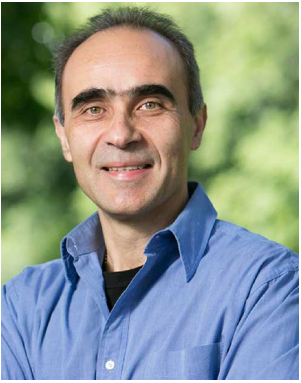 **George Bebis** received the B.S. degree in mathematics and the M.S. degree in computer science from the University of Crete, Crete, Greece, in 1987 and 1991, respectively, and the Ph.D. degree in electrical and computer engineering from the University of Central Florida, Orlando, FL, USA, in 1996. He is currently a Foundation Professor with the Department of Computer Science and Engineering (CSE), University of Nevada, Reno (UNR), Reno, NV, USA, and the Director of the Computer Vision Laboratory. From 2013 to 2018, he served as a Department Chair of CSE, UNR. His research has been funded by NSF, NASA, ONR, NIJ, and Ford Motor Company. His research interests include computer vision, image processing, pattern recognition, machine/deep learning, and evolutionary computing. Dr. Bebis is an Associate Editor of the Machine Vision and Applications Journal and serves on the Editorial Board of the International Journal on Artificial Intelligence Tools, and the Computer Methods in Biomechanics and Biomedical Engineering: Imaging and Visualization. He has served on the program committees of various national and international conferences and is the founder and steering committee chair of the International Symposium on Visual Computing (ISVC).

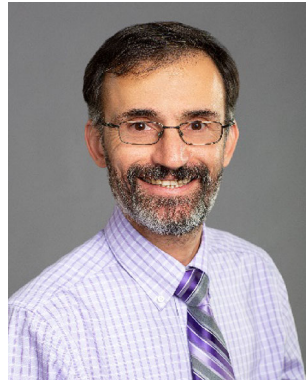 **Alireza Tavakkoli** is an Associate Professor in the Department of Computer Science and Engineering at the University of Nevada, Reno. He received his BSc and MSc degrees in Electrical Engineering from the Sharif University of Technology in 2001 and 2004, and MSC and PhD degrees in Computer Science from the University of Nevada, Reno in 2006 and 2009. He is the Director of the Human Machine Perception Lab at UNR. His main interests are in visual computing, artificial intelligence, and perception. His research projects are funded by federal agencies such as NSF, NASA, NIH and DoD. He has published over 100 peer-reviewed articles and occasionally serves as a grant review panelist as well as a reviewer for several journals and conferences. He is a senior member of the IEEE and currently the chief guest editor of a special research topic in the journal frontiers in ophthalmology.